\documentclass[10pt,twocolumn,letterpaper]{article}

\usepackage{cvpr}              

\usepackage[dvipsnames]{xcolor}

\definecolor{cvprblue}{rgb}{0.21,0.49,0.74}
\definecolor{cvprblueplusplus}{rgb}{0.1, 0.3, 0.95}
\definecolor{cvprred}{rgb}{0.85, 0.15, 0.15}

\usepackage[pagebackref,breaklinks,colorlinks,citecolor=cvprblue]{hyperref}

\usepackage{times}
\usepackage{epsfig}
\usepackage{graphicx}
\usepackage{amsmath}
\usepackage{amssymb}
\usepackage{booktabs}
\usepackage{multirow}
\usepackage{tabularx}
\usepackage{enumitem}
\usepackage{gensymb}
\usepackage{colortbl}
\usepackage{caption}
\usepackage{makecell}
\usepackage{wrapfig}
\usepackage{amssymb}
\usepackage{pifont}
\definecolor{mycyan}{cmyk}{.1,0,0,0}
\newcommand{\mypara}[1]{\vspace{1mm}\noindent\textbf{#1}}
\usepackage[capitalize]{cleveref}
\newcommand{\name}{AlignMiF}
\newcommand{\Geofusion}{GAA}
\newcommand{\Geoinit}{SGI}
\newcommand{\UniSim}{UniSim-SF}

\newcommand{\tr}[2][\@empty]{%
  \ifx#1\@empty
    \textcolor{red}{#2}%
  \else
    \textcolor{red}{\fontsize{#1}{#1}\selectfont #2}%
  \fi
}
\newcommand{\tb}[2][\@empty]{%
  \ifx#1\@empty
    \textcolor{blue}{#2}%
  \else
    \textcolor{blue}{\fontsize{#1}{#1}\selectfont #2}%
  \fi
}
\newcommand{\ts}[2][\@empty]{%
  \ifx#1\@empty
    {#2}%
  \else
    {\fontsize{#1}{#1}\selectfont #2}%
  \fi
}
\newcommand{\minus}{%
  \textcolor{cvprred}{\scalebox{0.92}[0.92]{(\texttt{-})}}%
}

\newcommand{\plus}{%
  \textcolor{cvprblueplusplus}{\scalebox{0.92}[0.92]{(\texttt{+})}}%
}

\usepackage{colortbl}
\definecolor{mygray}{gray}{.9}
\definecolor{mypink}{rgb}{.99,.91,.95}
\definecolor{mycyan}{cmyk}{.1,0,0,0}

\definecolor{F7E0D5}{RGB}{245,240,255}
\colorlet{Light}{White!0!F7E0D5}
\newcommand{\CC}{\cellcolor{Light}}
\newcommand{\LC}{\rowcolor{Light}}

\def\paperID{6883} 
\def\confName{CVPR}
\def\confYear{2024}

\title{\name{}: Geometry-Aligned Multimodal Implicit Field for \\ LiDAR-Camera Joint Synthesis}

\author{
  Tang Tao$^{1}$ \quad
  Guangrun Wang$^{3}$ \quad
  Yixing Lao$^{4}$ \quad
  Peng Chen$^{5}$ \quad 
  Jie Liu$^{6}$ \quad 
  Liang Lin$^{7}$ \quad 
  Kaicheng Yu$^{8}$ \footnotemark[2]\quad \\
  Xiaodan Liang$^{1,2}$ \footnotemark[2] \quad
  \\
\fontsize{11pt}{\baselineskip}\selectfont
$^1$ Shenzhen Campus of Sun Yat-sen University
\fontsize{11pt}{\baselineskip}\selectfont
$^2$ DarkMatter AI Research
\fontsize{11pt}{\baselineskip}\selectfont
$^3$ University of Oxford \\
\fontsize{11pt}{\baselineskip}\selectfont
$^4$ HKU
\fontsize{11pt}{\baselineskip}\selectfont
$^5$ Cainiao Group
\fontsize{11pt}{\baselineskip}\selectfont
$^6$ NCUT
\fontsize{11pt}{\baselineskip}\selectfont
$^7$ Sun Yat-sen University
\fontsize{11pt}{\baselineskip}\selectfont
$^8$ Westlake University \\
 \begin{normalsize}${\tt \{trent.tangtao, kaicheng.yu.yt,hxdliang328\}@gmail.com} $\end{normalsize}
}

\begin{document}
\maketitle
\renewcommand{\thefootnote}{\fnsymbol{footnote}}
\footnotetext[2]{Co-corresponding author.}
\renewcommand{\thefootnote}{\arabic{footnote}}
\begin{abstract}
Neural implicit fields have been a de facto standard in novel view synthesis. 
Recently, there exist some methods exploring fusing multiple modalities within a single field, aiming to share implicit features from different modalities to enhance reconstruction performance. However, these modalities often exhibit misaligned behaviors: optimizing for one modality, such as LiDAR, can adversely affect another, like camera performance, and vice versa. 
In this work, we conduct comprehensive analyses on the multimodal implicit field of LiDAR-camera joint synthesis, revealing the underlying issue lies in the misalignment of different sensors.
Furthermore, we introduce AlignMiF, a geometrically aligned multimodal implicit field with two proposed modules: Geometry-Aware Alignment (\Geofusion) and Shared Geometry Initialization (\Geoinit{}). These modules effectively align the coarse geometry across different modalities, significantly enhancing the fusion process between LiDAR and camera data. 
Through extensive experiments across various datasets and scenes, we demonstrate the effectiveness of our approach in facilitating better interaction between LiDAR and camera modalities within a unified neural field. Specifically, our proposed \name{}, achieves remarkable improvement over recent implicit fusion methods (\texttt{+}2.01 and \texttt{+}3.11 image PSNR on the KITTI-360 and Waymo datasets) and consistently surpasses single modality performance (13.8\% and 14.2\% reduction in LiDAR Chamfer Distance on the respective datasets). Code release: \href{https://github.com/tangtaogo/alignmif}{https://github.com/tangtaogo/alignmif}.

\end{abstract}

\vspace{-4pt}
\section{Introduction}
\label{sec:intro}
Synthesizing novel views has recently seen significant progress due to Neural Radiance Field (NeRF)~\cite{mildenhall2021nerf}, which models a 3D scene as a continuous function and leverages differentiable rendering, resulting in a de facto standard to render novel views.
Notably, the recent NeRF methods have shown impressive performance on downstream tasks such as autonomous diving~\cite{xie2023s-nerf, yang2023emernerf, wu2023mars, turki2023suds}.
In such practical scenarios, both images and LiDAR sensors are typically utilized. Currently, researchers extend the NeRF formulation for novel LiDAR view synthesis~\cite{tao2023lidarnerf, zhang2023nerflidar, huang2023neural}, which treat the oriented LiDAR laser beams as a set of rays and render 3D points and intensities in a similar fashion as RGB.


\begin{figure}[t]
    \centering
    \includegraphics[width=1\linewidth]{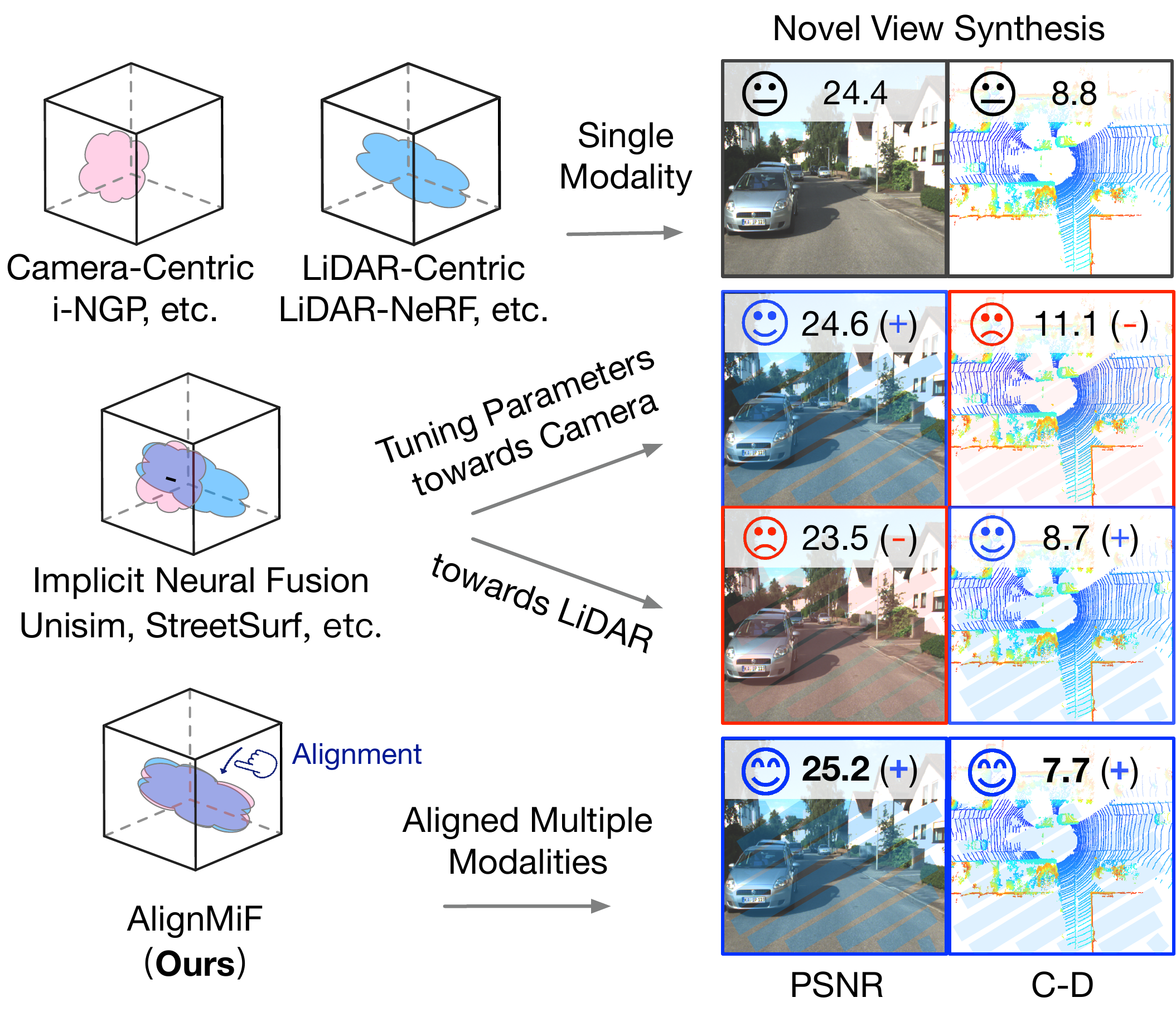}
     \vspace{-19pt}
    \caption{\textbf{The misalignment issue in multimodal implicit field.}
    For implicit neural fusion, there is a trade-off between the modalities due to the misalignment, making it challenging to improve both modalities simultaneously. Conversely, our method addresses the misalignment issue and achieves boosted multimodal performance. The metrics are PSNR and Chamfer Distance (C-D).
    }
    \label{fig:teaser}
    \vspace{-15pt}
\end{figure}

However, the exploration of multimodal learning in NeRF is still in its early stages. There are initial attempts, such as UniSim~\cite{yang2023unisim} and NeRF-LiDAR~\cite{zhang2023nerflidar}, to incorporate multimodal inputs through implicit fusion,  i.e., sharing the implicit features in one single field, aiming to leverage the complementary information from different modalities to enhance NeRF's capabilities. Accordingly, the integration of multiple input modalities is expected to boost model performance, but our results show that the naive multimodal NeRF, which relies on direct implicit fusion, does not outperform its unimodal counterpart.
As illustrated in \cref{fig:teaser}, it is challenging to improve both modalities simultaneously.
Intuitively, for NeRF optimization, incorporating more information is expected to lead to better results~\cite{wang2023digging, deng2022dsnerf, roessle2022depthpriors, yu2022monosdf, lao2023corresnerf}, which is not fully realized in current multimodal fields when fusing LiDAR and camera modalities. These observations highlight the need for ongoing research and advancements in multimodal learning in NeRF.

In this study, we perform comprehensive analyses of multimodal NeRF that integrates LiDAR and camera sensors for joint synthesis. Our preliminary experiments, conducted on real-world datasets such as KITTI-360~\cite{liao2022kitti360} and Waymo~\cite{sun2020waymo},
reveal that different modalities often contradict each other.
However, such conflicts are not observed in the synthetic AIODrive dataset~\cite{weng2023aiodrive}. These findings lead us to speculate that the underlying issue lies in the misalignment of different modalities, e.g., spatial misalignment and temporal misalignment. When modalities are not properly aligned, the implicit fusion of conflicting information can hinder network optimization, resulting in suboptimal outcomes for both modalities. To further investigate and validate this misalignment issue, we conducted extensive analyses, including the examination and visualization of raw sensor inputs, hash grid features, and the density values from the geometry network. Moreover, we conducted experimental analyses on various network architecture designs to validate our findings. These investigations provide valuable insights into the misalignment issue and its implications on the performance of the unified multimodal NeRF.

To tackle the challenge of misalignment in multimodal NeRF, we propose a twofold solution, called \name{}, to align the consistent coarse geometry across different modalities while keeping their individual detail characteristics. Firstly, we decompose the hash encoding to allow each modality to concentrate on its own information. We then apply an alignment constraint at the coarse geometry levels, facilitating mutual enhancement and cooperation between modalities, referred to as the Geometry-Aware Alignment (\Geofusion{}) module.
Secondly, we utilize the hash grid features from a pre-trained field as a share initialization of the geometry, referred to as the Shared Geometry Initialization (\Geoinit{}) module. This shared initialization further enhances the alignment process, allowing each modality to capture its respective details upon it. Both \Geofusion{} and \Geoinit{} aim to align the coarse geometry while preserving their unique details.

Through comprehensive experiments conducted on multiple datasets and scenes, we validate the effectiveness of our approach in boosting the interaction between LiDAR and camera modalities within a unified framework.
Our proposed modules, \Geofusion{} and \Geoinit{}, contribute to improved alignment and fusion, leading to enhanced performance and more accurate synthesis of novel views.
Specifically, as a result, our \name{} achieves remarkable improvement over the implicit fusion (e.g., \texttt{+}2.01 and \texttt{+}3.11 PSNR on KITTI-360 and Waymo datasets) and consistently outperforms the single modality (e.g., 13.8\% and 14.2\% reduction in LiDAR Chamfer Distance on the respective datasets).

Overall, our contributions are as follows:
\begin{itemize}
    \item We perform comprehensive analyses of multimodal learning in NeRF, identifying the modality misalignment issue.
    \item We propose \name{}, with \Geofusion{} and \Geoinit{} modules, to address the misalignment issue by aligning the consistent coarse geometry of different modalities while preserving their unique details.
    \item We demonstrate the effectiveness of our method quantitatively and qualitatively through extensive experiments conducted on multiple datasets and scenes.
\end{itemize}

\section{Related Work}
\label{sec:related}
\subsection{NeRF for Novel View Synthesis}
Neural Radiance Fields (NeRF)~\cite{mildenhall2021nerf} have revolutionized the long-standing novel view synthesis. Many NeRF variants have been proposed, focusing on aspects such as acceleration~\cite{muller2022instantNGP, yu2021plenoxels, chen2022tensorf}, anti-aliasing~\cite{barron2021mip, barron2023zip, hu2023tri}, managing casual camera trajectories~\cite{wang2023f2, meuleman2023progressively}, and generalization capabilities~\cite{yu2021pixelnerf, huang2023local}. There also emerges research leveraging depth information for view synthesis~\cite{deng2022dsnerf, neff2021donerf, roessle2022depthpriors, wang2023digging}. In parallel, great progress has been made in NeRF applications for handling complex and large-scale environments such as urban outdoor scenes~\cite{rematas2022urban-nerf, tancik2022blocknerf, xie2023s-nerf, wu2023mars, lu2023dnmp, yang2023urbangiraffe, li2023read, turki2023suds, yang2023emernerf}.
Concurrently, researchers extend the NeRF formulation for novel LiDAR view synthesis~\cite{tao2023lidarnerf, zhang2023nerflidar, huang2023neural, zheng2023neuralpci, hu2023pcnerf}, which treat the oriented LiDAR laser beams in a similar manner to camera rays. Given the recent advancements for novel view synthesis in different modalities, in this work, we dig into the investigation of the unified multimodal NeRF framework.

\subsection{Multimodal Learning in NeRF}
Recent works have also explored multi-task learning in NeRF, which involves synthesizing panoptic or semantic labels alongside RGB views~\cite{fu2022panoptic, zhi2021semanticnerf, zheng2023multi}.
However, the exploration of multimodal learning in NeRF is still in its early stages. Some current works have attempted to incorporate multimodal inputs, such as NeRF-LiDAR~\cite{zhang2023nerflidar}, which leverages images, semantic labels, and LiDAR data to generate LiDAR points and corresponding labels.
Another work, StreetSurf~\cite{guo2023streetsurf}, utilizes LiDAR as supervision for street view multi-view reconstruction and can also generate 3D points.
UniSim~\cite{yang2023unisim}, on the other hand, is a neural sensor simulator that takes multi-sensor inputs into a shared implicit field,
and simulates LiDAR and camera data at new viewpoints.
These preliminary efforts primarily focus on simple implicit fusion, i.e., directly sharing the implicit features of different modalities in a single field. However, we actually find that modalities in this multimodal NeRF often contradict each other and cannot outperform its unimodal counterpart. Our work delves deeper into exploring the intricate interactions between multimodalities and proposes a geometry-aligned multimodal implicit field.

\section{Problem Analysis}
\label{sec:problem}

\subsection{Preliminaries}
NeRF~\cite{mildenhall2021nerf} models a scene as a continuous volumetric field.
Given a 3D location $\mathbf{x}$ and a
viewing direction $\boldsymbol{\theta}$ as input, NeRF learns an implicit function $f$ that predicts the volume density $\sigma$ and color $\mathbf{c}$ as $(\sigma, \mathbf{c}) = f(\mathbf{x}, \boldsymbol{\theta})$.
Specifically, given rays $\mathbf{r}$ originated from camera origin $\mathbf{o}$ in direction $\mathbf{d}$, i.e., $\mathbf{r}(t) = \mathbf{o} + t\mathbf{d}$, the corresponding pixel color is approximated by the numerical quadrature of the color $\mathbf{c}_i$ and density $\sigma_i$ of samples along the ray: $\hat{C}(\mathbf{r}) = \sum_{i=1}^N T_i \big(1 - \exp(-\sigma_i \delta_i)\big)\mathbf{c}_i
\label{eq:nerf}$,
where $T_i = \exp (-\sum_{j=1}^{i-1} \sigma_j \delta_j )$ and $\delta_i$ is the distance between adjacent samples.

More recent works~\cite{tao2023lidarnerf, zhang2023nerflidar, huang2023neural} extend the traditional NeRF to LiDAR sensor, treating the oriented LiDAR laser beams as a set of rays.
Slightly abusing the notation, let $\mathbf{r}(t) = \mathbf{o} + t\mathbf{d}$ be a ray casted from the LiDAR sensor, where $\mathbf{o}$ denotes the LiDAR center, and $\mathbf{d}$ represents the normalized direction vector of the corresponding beam.
Then the depth measurement $\hat{D}(\mathbf{r})$ can be approximated by calculating the expectation of the samples along the ray: $\hat{D}(\textbf{r}) = \sum_{i=1}^N T_i \big(1 - \exp(-\sigma_i \delta_i)\big)t_i \label{eq:depth}$.
The view-dependent features of LiDAR, including the intensities and ray-drop probabilities, can be rendered similarly to RGB color.

\subsection{Multimodal Learning in NeRF}

\begin{table}[ht]
  \centering \small
\caption{\textbf{The misalignment issue.} When directly implicit fusing the multiple modalities cannot outperform their unimodal counterparts simultaneously in real-world datasets. While this challenge is not present in the synthetic dataset. Here, $w$ denotes the weight ratio between the two modalities, $w_\lambda = \lambda_c / \lambda_l$.}
\vspace{-2pt}
  \addtolength{\tabcolsep}{-3.1pt}
    \begin{tabularx}{\linewidth}{l|c|cc|cc}
  \toprule
 \multirow{2}{*}{Method}   &  \multirow{2}{*}{$w_\lambda$}  & \multicolumn{2}{c|}{RGB Metric} & \multicolumn{2}{c}{LiDAR Metric}  \\
      \cmidrule(r){3-6} &   & PSNR$\uparrow$ & SSIM$\uparrow$  & C-D$\downarrow$ & F-score$\uparrow$ \\
      \midrule
       \multicolumn{3}{l}{\textit{\textbf{KITTI-360} \ts[9pt]{(real-world)}}}    \\
        \midrule

    i-NGP~\cite{muller2022instantNGP} &--& 24.45 &0.787 & --&--\\
     LiDAR-NeRF~\cite{tao2023lidarnerf} & --&--&--&0.088 & 0.920\\
     \midrule
        \multirow{3}{*}{\UniSim ~\cite{yang2023unisim}}
         & 0.1& 23.54 \minus  &0.759  &0.087 \plus& 0.929 \\
                & 0.5& 24.38 \minus& 0.792 &0.100 \minus & 0.920\\
               & 2.0 &24.80 \plus & 0.809 &0.124 \minus &0.900\\

     \midrule
       \multicolumn{3}{l}{\textit{\textbf{Waymo} \ts[9pt]{(real-world)}}}    \\
        \midrule
    i-NGP~\cite{muller2022instantNGP} &--& 28.20 &0.830 & --&--\\
     LiDAR-NeRF~\cite{tao2023lidarnerf} & --&--&--&0.179 & 0.885\\
     \midrule
       \multirow{3}{*}{\UniSim ~\cite{yang2023unisim}}
                & 0.5& 26.41 \minus & 0.789 & 0.172 \plus & 0.891\\
    & 1.0 & 27.12 \minus &0.805& 0.181 \minus & 0.885\\
                & 5.0 &28.33 \plus & 0.830 &0.227 \minus & 0.840\\

    \midrule
       \multicolumn{3}{l}{\textit{\textbf{AIODrive} \ts[9pt]{(synthetic)}}}    \\
        \midrule
         i-NGP~\cite{muller2022instantNGP} &--& 34.43 &0.893 & --&--\\
     LiDAR-NeRF~\cite{tao2023lidarnerf} & --&--&--&0.178 & 0.873\\
     \midrule
       \multirow{3}{*}{\UniSim ~\cite{yang2023unisim}}
       & 0.1& 34.25  \minus &  0.901 & 0.138 \plus& 0.921\\
      & 1.0 &  \CC 34.53 \plus & \CC 0.904 & \CC 0.153 \plus & \CC 0.915\\
                & 5.0 &34.64 \plus & 0.905 &0.191 \minus &0.914\\

      \bottomrule
    \end{tabularx}
  \label{tab:aiodrive}
  \vspace{-10pt}
\end{table}

Building on the NeRF formulations for different sensors mentioned above, there have been preliminary works, such as UniSim~\cite{yang2023unisim} and NeRF-LiDAR~\cite{zhang2023nerflidar}, integrating these into a unified multimodal NeRF framework. These methods directly share implicit features across different modalities, and the optimization targets can be combined as:
 \begin{equation}
      \mathcal{L}_{\mathrm{total}} =
      \lambda_l\mathcal{L}_{\mathrm{LiDAR}}(\mathbf{r_l})  +
      \lambda_c \mathcal{L}_{\mathrm{camera}}(\mathbf{r_c}) ,
    \label{eq:loss}
\end{equation}
where $\mathbf{r_l}\in R_l$ and $\mathbf{r_c}\in R_c$
are the sensor training rays,
and $\lambda$ are weight coefficients to balance each term.
However, current efforts only primarily focus on simple implicit fusion, and the multimodal NeRF has not been fully exploited.

\subsection{The Misalignment Issue}
\label{subsec:misalign}
Multimodal learning helps to comprehensively understand the world, by integrating different senses. Accordingly, multiple input modalities are expected to boost model performance. However, our findings suggest that the current multimodal NeRF cannot surpass its uni-modal counterpart in terms of performance.
As shown in \cref{tab:aiodrive},
regardless of how we adjust the weights of the two modalities, we find that it is challenging to improve both modalities simultaneously (more tuning results are shown in \cref{fig:teaser_old} of the appendix).
Indeed, similar challenges have also been observed in other multimodal research areas. Previous researchers claimed that different modalities exhibit inconsistent representations and tend to converge at different rates, leading to uncoordinated convergence problems~\cite{peng2022balanced, sun2021learning, wang2020makes}. However, these theories are not applicable to our scenario, as our representation is a unified field that reflects the real world.
Intuitively, for NeRF optimization, it is expected that having more information would lead to better results~\cite{deng2022dsnerf, neff2021donerf, roessle2022depthpriors, yu2022monosdf, lao2023corresnerf}.
We further conducted experiments on the synthetic dataset, AIODrive~\cite{weng2023aiodrive}, which was collected from CARLA Simulator~\cite{dosovitskiy2017carla}.
As shown in the bottom block of \cref{tab:aiodrive}, we can observe that mutual boosting between the two modalities could be achieved on the perfect synthetic data. Based on these observations, we speculate that the underlying issue lies in the misalignment of modalities.
When two modalities are not properly aligned, the implicit fusion of conflicting information causes the network to struggle to determine the correct optimization direction, resulting in suboptimal results for both.


\begin{figure}[ht]
    \centering
    \includegraphics[width=1\linewidth]{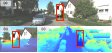}
    \vspace{-14pt}
    \caption{\textbf{Analysis of misalignment from raw sensor inputs.} (a) Original image, (b) Image with projected points from associate LiDAR frame, (c) LiDAR points of all scene frames. As highlighted in the red box, the observations obtained from LiDAR and camera sensors for the same pole are distinct (zoom-in for better views).}
    
    \label{fig:migalign_sensor}
    \vspace{-13pt}
\end{figure}

To further investigate the misalignment issue, we begin with the raw modality inputs.
As illustrated in \cref{fig:migalign_sensor}, both the camera and LiDAR sensors essentially represent the same overall scene, yet they exhibit variances in capturing finer details.
For example, when scanning the same lamp post, the pole obtained from LiDAR appears to have larger diameter compared to the one captured by the RGB camera.
In fact, the perceptual characteristics of these sensors are inherently different.
LiDAR lacks semantic perception of objects and provides rougher boundaries, while the camera lacks distance perception.
Moreover, even without considering calibration errors between multiple sensors, inherent systematic errors exist in each sensor~\cite{yu2022benchmarking, cao2022Misalignment}, e.g., the different operating frequency and trigger mechanism of camera, LiDAR, GPS, and IMU.
Consequently, it becomes challenging to align all details across the two modalities, resulting in ambiguous conflicts within one unified field.

\begin{figure}[t]
    \centering
    \includegraphics[width=1\linewidth]{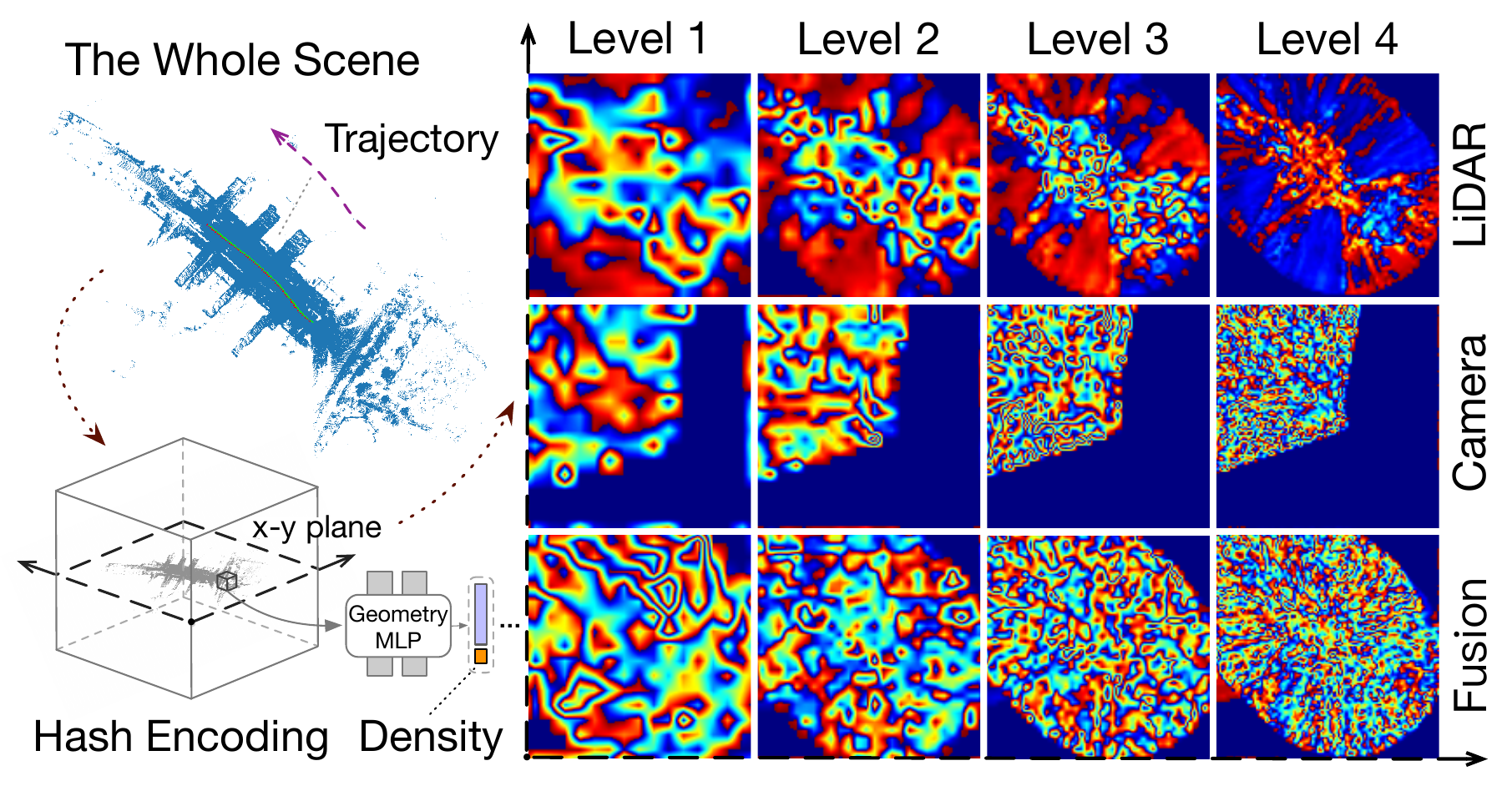}
    \vspace{-18pt}
    \caption{\textbf{Analysis of misalignment from bird's eye view hash grid features.} We show the first 4 levels of the hash features on the x-y plane. The camera is front-facing along the trajectory and brighter or more saturated colors represent higher feature values.
    }
    \label{fig:misalign_feat}
    \vspace{-9pt}
\end{figure}

Next, we investigate the learned hash gird features of different modalities.
The multi-resolution hash encoding introduced by iNGP~\cite{muller2022instantNGP}
is expressive and efficient, which is a common practice in current works~\cite{tao2023lidarnerf, yang2023unisim, guo2023streetsurf}.
In \cref{fig:misalign_feat}, we visualize the learned hash features on the x-y plane.
It is apparent that the learned geometry from LiDAR is well represented, and the shape of the hash features aligns with the scene. Then, both modalities primarily focus on their respective field of view  (FOV), with the camera capturing the top-left part due to its front-facing orientation. Notably, even in the overlap area of the FOV, the highlighted feature regions of interest differ, indicating conflicting ambiguity between the two modalities.
Consequently, when utilizing the simple implicit fusion, i.e., directly sharing the implicit features of different modalities, the model becomes confused by the misaligned modalities, resulting in disorganized hash features (bottom row in \cref{fig:misalign_feat}), ultimately yielding suboptimal results for both.

Furthermore, we present additional analysis and experiments in \cref{fig:geo_fusion} and \cref{tab:different_arch}. These findings both provide valuable evidence to support our claims and contribute
to a deeper understanding of the misalignment issue.

\section{\name}
\label{sec: mmnerf}
In this section, we introduce our multimodal implicit field, \name{}, with two proposed geometrical alignment modules, aiming to mitigate the misalignment problem and enhance performance across both modalities.
\begin{figure*}[h]
    \centering
    \includegraphics[width=1\linewidth]{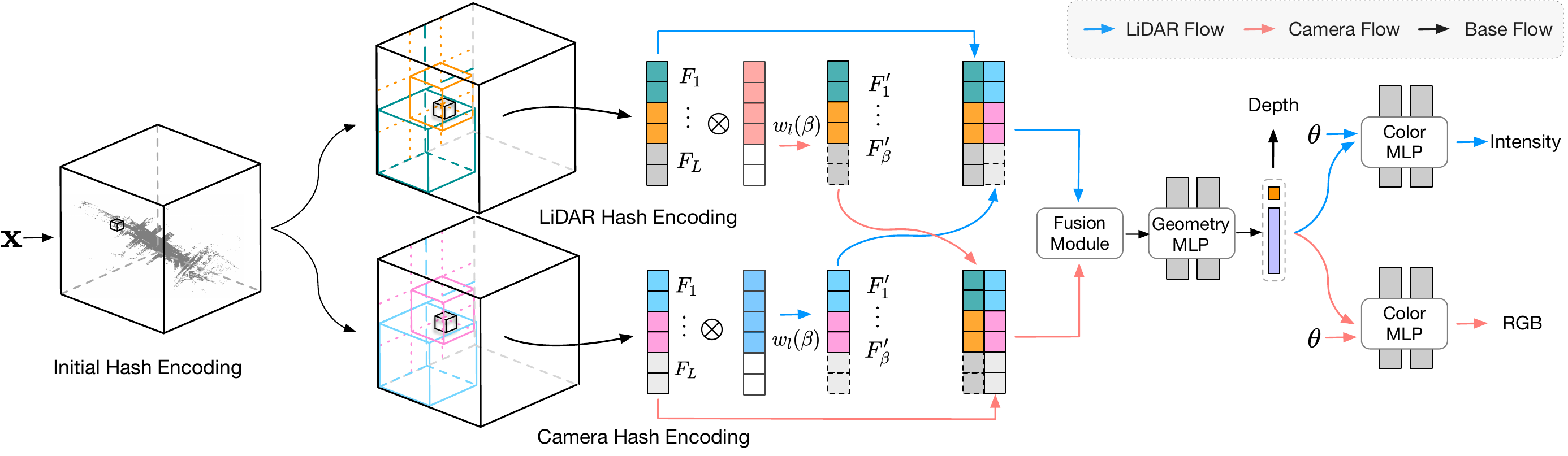}
    \vspace{-15pt}
    \caption{\textbf{The illustration of our \name{} framework.}  The proposed Geometry-Aware Alignment (\Geofusion{}) of the decomposed hash encoding and the Shared Geometry Initialization (\Geoinit{}) are incorporated together to tackle the misalignment issue.}
    \label{fig:mmnerf}
    \vspace{-8pt}
\end{figure*}

\subsection{Geometry-Aware Alignment}
\label{subsec:geo_fusion}

To alleviate the misalignment issue between the two modalities, we first decompose the hash encoding, allowing each modality to focus on its own information; while subsequently, we need to enhance information interactions between modalities.
Inspired by Neuralangelo~\cite{li2023neuralangelo} and HR-Neus~\cite{liang2023hr}, we acknowledge that different levels of hash encoding correspond to different levels of fine-grained geometry information. Specifically, the lower-indexed grid levels contain the most information about the coarse geometry, while the higher-level grids primarily contain information about high-frequency details.
As observed in our earlier analysis, the misalignment primarily occurs at the detailed levels, while both modalities share the same underlying coarse scene geometry of the real world.
Building upon this observation, we further propose a \textbf{G}eometry-\textbf{A}ware \textbf{A}lignment (GAA), which specifically aligns the two modalities at the coarse geometry levels, facilitating mutual enhancement and cooperation between them.

Specifically, we denote the multi-resolution hash encoding with total $L$ levels and a feature dimension of $d$ as $\gamma$. Given a query point $\mathbf{x}$, the 3D feature grid at each level is first trilinearly interpolated and then concatenated together to form the final feature vector:
$
\gamma(\mathbf{x}) = \overline{F_1 F_2 \dots F_L} \in \mathbb{R}^{L \times d}$,
where $F_l \in \mathbb{R}^d$ represents the interpolated feature at level $l$. Then as previous works~\cite{li2023neuralangelo,liang2023hr}, we expand the definition of $\gamma$ to take in an additional parameter $\beta$  that helps zero out the higher-level resolution feature grid:
\begin{align}
    \gamma(\mathbf{x} , \beta) = \overline{F_1' F_2' \dots F_L'}, \ \text{where}  \ F_l'  = w_l(\beta) F_l,  \nonumber  \\
    w_l(\beta) = \frac{1 - \cos(\pi \cdot \text{clamp}(\beta - l + 1, 0, 1))}{2}
    \label{eq:multires}
\end{align}
Intuitively, grid layer $F_l$ will be fully activated if $l \leq \beta$ and all other higher grid layers will be zeroed out.  Note that all grid layers are available when $\beta$ is not provided.
We denote the two decomposed hash encoding as $\gamma_{lidar}$ and $\gamma_{camera}$, and then our GAA can be formulated as:
\begin{equation}
   \psi_{GAA}(\mathbf{x}, \beta) =
   \begin{cases}
  \mathcal{F} (\gamma_{lidar}(\mathbf{x}), \gamma_{camera}(\mathbf{x} , \beta) ) & \mathbf{x} \sim \mathbf{r_l} \\
  \mathcal{F} (\gamma_{lidar}(\mathbf{x}, \beta), \gamma_{camera}(\mathbf{x} ) ) & \mathbf{x} \sim \mathbf{r_c}\\
   \end{cases},
    \label{eq:gaa}
\end{equation}
where $\mathcal{F}$ denotes the fusion module for the alignment, e.g., concatenation or attention mechanism. Shortly, our GAA combines the hash features of the current modality with the aligned coarse level hash features from another modality and the illustration is on \cref{fig:mmnerf}.

\begin{figure}[t]
    \centering
    \includegraphics[width=1\linewidth]{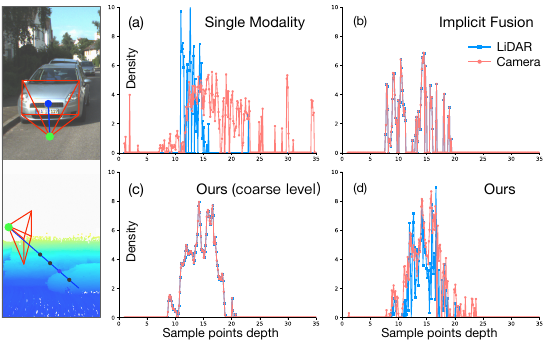}
    \vspace{-20pt}
    \caption{\textbf{Analysis of misalignment from the density values and qualitative analysis of our propose \Geofusion.}} 
    \label{fig:geo_fusion}
    \vspace{-4pt}
\end{figure}

In \cref{fig:geo_fusion}, we visualize the learned density values from the geometry MLP for sampled points along the ray.
In (a), the density values, reflecting the geometry, learned by the LiDAR and camera are very inconsistent, which correspond to the previously obtained hash features in \cref{fig:misalign_feat}.
In (b), when employing the implicit fusion, the learned densities can be confounded by misalignment between the two modalities, resulting in erroneous geometry.
Conversely, in (c) and (d), our method achieves a more robust and comprehensive geometry representation that incorporates information from both modalities. Specifically, the GAA module aligns the coarse level of geometry, i.e., the consistent densities, as shown in (c), while both modalities capture their respective finer details as in (d), which demonstrates the effectiveness of our method.

\subsection{Shared Geometry Initialization}
\label{subsec:geo_init}
\begin{figure}[h]
    \centering
    \includegraphics[width=1\linewidth]{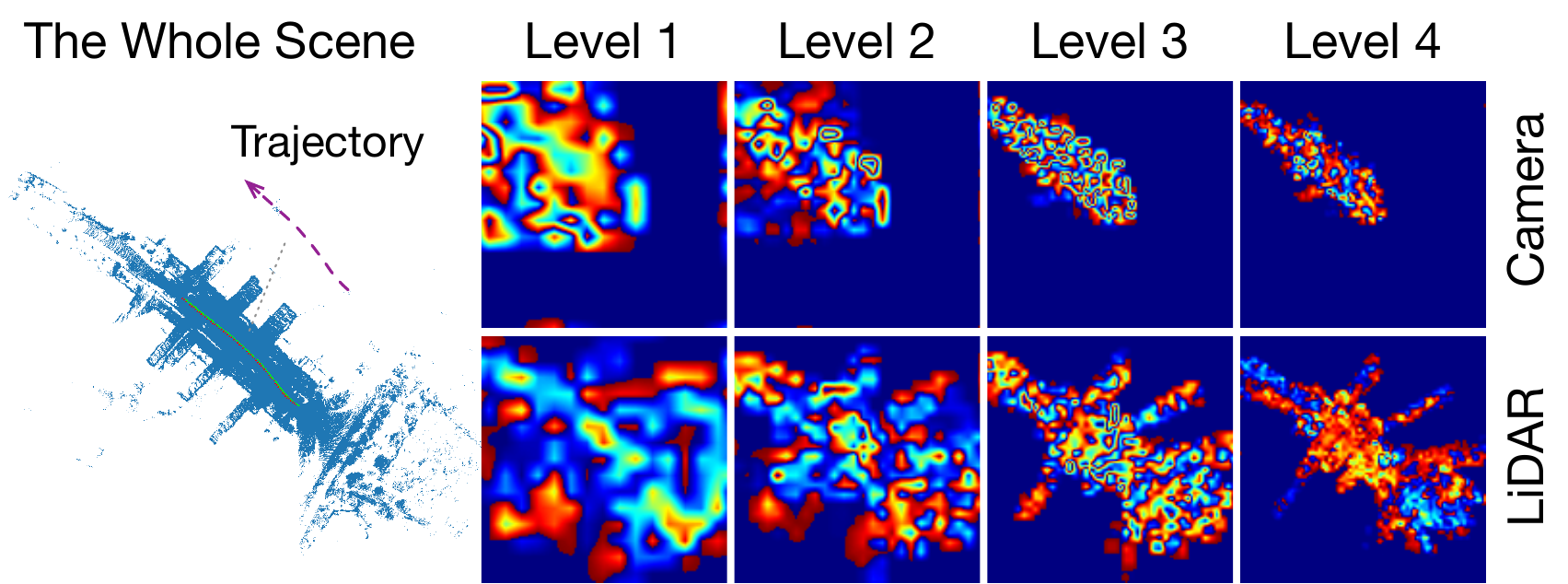}
    \vspace{-15pt}
    \caption{\textbf{Qualitative analysis of our propose \Geoinit{} module.}}
    \label{fig:geo_init}
    \vspace{-10pt}
\end{figure}

In the visualizations of the hash features and density values shown in \cref{fig:misalign_feat} and \cref{fig:geo_fusion}, it is evident that the learned geometry from the camera can be inaccurate. This observation also conformed with the shape-radiance ambiguity discussed in previous works~\cite{neff2021donerf, wang2023digging, deng2022dsnerf}. Although we propose a geometry-aware alignment module to enhance the camera's geometry using LiDAR information, the learning and alignment process still remain implicit.
To address this, we propose utilizing the hash grids from a pre-trained field with rough geometry, e.g., the trained LiDAR field, as a shared geometry initialization for both modalities. Previous works~\cite{yang2023unisim, carlson2023cloner} have also suggested using LiDAR to constrain the volume grids. However, considering the different FOV and the varying details of LiDAR and camera data as analyzed earlier, we additionally learn the hash features of both modalities after the initialized hash encoder, rather than relying only on LiDAR.
Then we directly add hash features from the initial encoding to each modality as the shared geometry information.
Specifically, we denote the shared initialized hash grid as $\gamma_{init}$, and the proposed \textbf{S}hared \textbf{G}eometry  \textbf{I}nitialization (\Geoinit{}) can be formulated as:
\vspace{-4pt}
\begin{equation}
   \phi_{\Geoinit{}}(\mathbf{x}) =
   \begin{cases}
 \gamma_{init}(\mathbf{x}) + \gamma_{lidar}(\mathbf{x})  & \mathbf{x} \sim \mathbf{r_l} \\
  \gamma_{init}(\mathbf{x}) + \gamma_{camera}(\mathbf{x}) & \mathbf{x} \sim \mathbf{r_c}\\
   \end{cases} .
    \label{eq:\Geoinit{}}
\end{equation}
As shown in \cref{fig:geo_init}, after applying our \Geoinit{} module, we observe remarkable improvements in the hash features of both LiDAR and camera encoding, compared with \cref{fig:misalign_feat}. Especially, the camera's hash features exactly focus on the relevant regions of the scene geometry.
These observations demonstrate the effectiveness of our proposed module.

In summary, our \Geoinit{} leverages the initialization from a pre-trained field as the shared coarse geometry while still allowing the learning of hash features from both modalities to capture their respective details. This aligns with the GAA module, and both of them align coarse geometry while preserving modalities' unique characteristics.


\subsection{\name{} Formulation}
Overall, as illustrated in \cref{fig:mmnerf}, combining the proposed two simple yet effective modules, we summarize the formulation of our \name{} as follows:
\begin{align}
   \Psi&_{GAA}^{\Geoinit{}}(\mathbf{x}, \beta) =  \psi_{GAA}(\mathbf{x}, \beta;
   \;\phi_{\Geoinit{}}) \nonumber \\
   & =
   \begin{cases}
  \mathcal{F} (
  \phi_{lidar}(\mathbf{x})), \ \phi_{camera}(\mathbf{x} , \beta) ) & \mathbf{x} \sim \mathbf{r_l} \\
  \mathcal{F} (\phi_{lidar}(\mathbf{x}, \beta), \  \phi_{camera}(\mathbf{x}) ) & \mathbf{x} \sim \mathbf{r_c}\\
   \end{cases},
    \label{eq:mmnerf}
\end{align}
\vspace{-4pt}
where $\phi_{\Geoinit{}}(\mathbf{x} , \beta) =  \gamma_{init}(\mathbf{x}) + \gamma(\mathbf{x}, \beta) $.

\section{Experiment}
\label{sec:exp}

\subsection{Experimental Setting}
\mypara{Datasets.}
We conducted experiments on three datasets: one synthetic dataset, AIODrive~\cite{weng2023aiodrive}, and two challenging real-world datasets, KITTI-360~\cite{liao2022kitti360} and Waymo Open Dataset~\cite{sun2020waymo}.
These datasets were collected using RGB cameras and LiDAR sensors.

\mypara{Evaluation metrics.}
For novel image view synthesis, following the previous methods~\cite{mildenhall2021nerf, muller2022instantNGP}, our evaluations are based on three widely-used metrics, \ie, peak signal-to-noise ratio (PSNR), structural similarity index measure (SSIM)~\cite{wang2004ssim}, and the learned perceptual image patch similarity (LPIPS)~\cite{zhang2018lpips}.
For novel LiDAR view synthesis, as works~\cite{tao2023lidarnerf, huang2023neural}, we report the Chamfer Distance (C-D) between the rendered and original LiDAR point clouds and the F-Score with a threshold of 5cm.
The novel intensity image is evaluated using mean absolute error (MAE).


\mypara{Baselines.}
For the uni-modal model, we consider the popular i-NGP~\cite{muller2022instantNGP} for novel image view synthesis and the concurrent LiDAR-NeRF~\cite{tao2023lidarnerf} for novel LiDAR view synthesis. For multimodal evaluation, we use UniSim~\cite{yang2023unisim} as the main baseline.
Since this work focuses on investigating the relationship between multimodalities, we specifically re-implement its implicit fusion component and also did not consider dynamic foreground. To distinguish it from the original UniSim, we refer to it as UniSim-Static-Implicit Fusion, abbreviated as UniSim-SF.
Details of dataset sequences and splits and implementation details are provided in supplementary materials.

\subsection{Main Results}
\begin{table*}[ht]
  \centering
  \small
  \caption{\textbf{Novel view synthesis on KITTI-360 dataset and Waymo dataset}. \name{} outperforms the baselines in all metrics.
  }
  \addtolength{\tabcolsep}{-4.65pt}
    \begin{tabularx}{0.95\linewidth}{l|c|ccc|ccc||ccc|ccc}
  \toprule
 \multirow{3}{*}{Method} &
 \multirow{3}{*}{M}  & \multicolumn{6}{c||}{KITTI-360 Dataset} & \multicolumn{6}{c}{Waymo Dataset}  \\
 \cmidrule(r){3-14} & & \multicolumn{3}{c|}{RGB Metric} & \multicolumn{3}{c||}{LiDAR Metric} & \multicolumn{3}{c|}{RGB Metric} & \multicolumn{3}{c}{LiDAR Metric} \\
       &    & PSNR$\uparrow$ & SSIM$\uparrow$ & LPIPS$\downarrow$ & C-D$\downarrow$ & F-score$\uparrow$  & MAE$\downarrow$ & PSNR$\uparrow$ & SSIM$\uparrow$ & LPIPS$\downarrow$ & C-D$\downarrow$ & F-score$\uparrow$  & MAE$\downarrow$\\
      \midrule

      i-NGP~\cite{muller2022instantNGP}&C & 24.61 & 0.808 & 0.181 & --&--&-- &28.82	& 0.831	& 0.380	& --&--&--\\
     LiDAR-NeRF~\cite{tao2023lidarnerf}&L & --&--&--& 0.094 & 0.916 & 0.122
     & --&--&--&	0.197 &	0.871 & 0.040 \\
     $\text{\UniSim ~\cite{yang2023unisim}}^{\vartriangle}$  &LC & 23.30 \minus  & 0.758 & 0.268 & 0.090 \plus& 0.924 & 0.097
     & 26.67 \minus 	&0.788	&0.417&	0.186 \plus&	0.878	&0.039 \\
     $\text{\UniSim ~\cite{yang2023unisim}}^{\triangledown}$   &LC & 24.94 \plus&	0.812&	0.184&	0.114 \minus&0.906&	0.095
     & 28.98 \plus&	0.833&	0.374 &	0.355 \minus&	0.786 &	0.045 \\
   \LC   \name{} (ours)  &LC & 25.31 \plus&	0.826	&0.164&	0.081 \plus&	0.928	&0.099 &
   29.78 \plus	&0.845 &	0.339&	0.169 \plus &	0.885&	0.038\\

      \bottomrule
      \multicolumn{14}{l}{\scriptsize{M, L, C denotes modality, LiDAR, camera respectively. $\vartriangle$ and $\triangledown$ represent tuning parameters towards LiDAR and camera modality respectively.}}
    \end{tabularx}
  \label{tab:all_res}
  \vspace{-15pt}
\end{table*}

\mypara{Results on KITTI-360 and Waymo dataset.}
In \cref{tab:all_res}, we present the evaluation results on the KITTI-360 and Waymo datasets. As mentioned in \cref{subsec:misalign}, the \UniSim{} model fails to achieve simultaneous improvements in multimodal performance, and there is a trade-off between the modalities. To ensure a fair comparison, we carefully fine-tuned the parameters to ensure that the LiDAR or camera modality in \UniSim{} surpassed its corresponding single-modality counterpart by a small margin, preventing either modality from being significantly worse.
Compared with the carefully tuned \UniSim{} model and the corresponding single-modality models, our \name{} achieves superior results by clear margins.
Specifically, our method achieves remarkable improvement over \UniSim{} by \texttt{+}2.01 and \texttt{+}3.11 image PSNR on KITTI-360 and Waymo datasets respectively, and outperforms the single-modality models overall, as evidenced by the 13.8\% and 14.2\% reduction in LiDAR Chamfer Distance respectively.
With two proposed alignment modules, our method facilitates better fusion of different modalities, resulting in more accurate understanding of the scene and improved results.

\begin{figure}[h]
    \centering
    \includegraphics[width=1\linewidth]{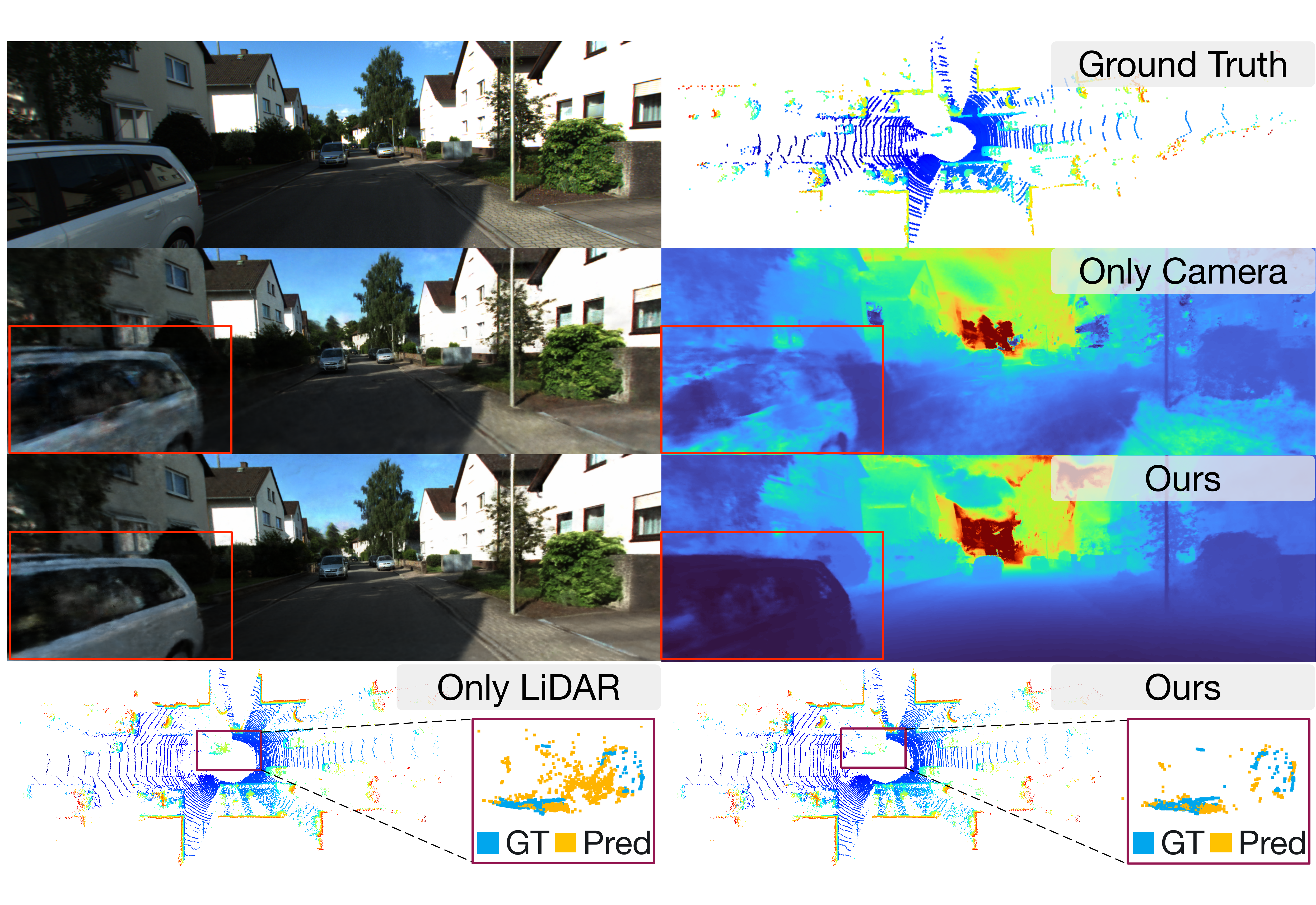}
    \vspace{-20pt}
    \caption{\textbf{Qualitative results on KITTI-360 dataset.} Our method achieves mutual boosting between both modalities (zoom-in for the best of views). 
    }
    \label{fig:res}
    \vspace{-15pt}
\end{figure}

Furthermore, we provide qualitative results in \cref{fig:res}, which clearly demonstrate the mutual benefits of our \name{}. As highlighted with the boxes, the LiDAR modality significantly enhances the learning of image and depth quality in the camera, while the semantic information from RGB aids the LiDAR in better converging to object boundaries.
Please refer to our supplementary materials for all scene results and more visualization results.

\begin{table}[ht]
  \centering \small
  \caption{\textbf{Comparison with StreetSurf on Waymo dataset.}
  }
   \addtolength{\tabcolsep}{2.5pt}
    \begin{tabularx}{\linewidth}{l|cc|cc}
  \toprule
 \multirow{2}{*}{Sequence}  & \multicolumn{2}{c|}{StreetSurf~\cite{guo2023streetsurf} } & \multicolumn{2}{c}{\name{}}  \\
       \cmidrule(r){2-5}   & C & LC & C & LC \\
      \midrule
        seg1137922...   & 28.33 &  27.64 & 29.26 &  30.16\\
        seg1067626...   & 29.01 &  27.68 & 29.52 &  30.27\\
        seg1776195...   & 26.71 &  25.35 & 28.20 &  29.22\\
        seg1172406...   & 28.50 &  27.86 & 28.30 &  29.47\\

       \midrule
       Average & 	28.14 & 	27.13 \textbf{\minus}& 28.82& 29.78 \textbf{\plus}\\

      \bottomrule
      \multicolumn{5}{l}{\scriptsize{L, C denotes LiDAR, camera respectively and the metric is PSNR$\uparrow$.}}
    \end{tabularx}
  \label{tab:sota}
  \vspace{-16pt}
\end{table}

\mypara{Results compared with StreetSurf.}
In \cref{tab:sota}, we specifically compare the PSNR metric with StreetSurf~\cite{guo2023streetsurf} as it does not learn LiDAR intensity and ray-drop.
It is worth noting that StreetSurf produces lower image results with the implicit LiDAR-camera fusion compared to the single-camera modality. This further emphasizes the misalignment issue across different representations and methods.
In contrast, our method successfully improves the performance of both modalities and achieves state-of-the-art results.

\subsection{Ablation Study}
\label{subsec:ablation}
\begin{figure}[t]
    \centering
    \includegraphics[width=1\linewidth]{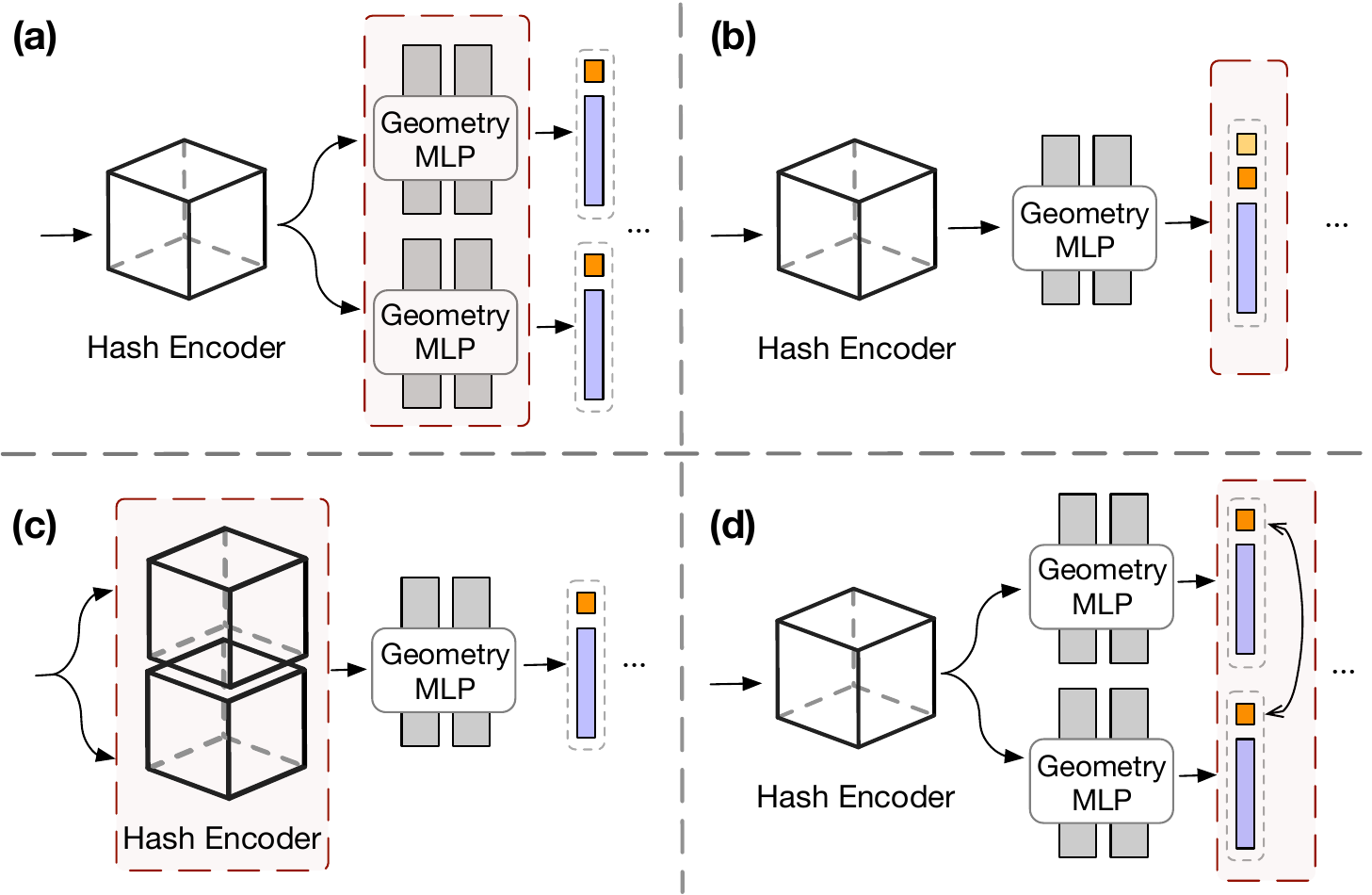}
    \vspace{-16pt}
    \caption{\textbf{Different designs of the network architecture.} (a) Decompose geometry net, (b) Decompose densities, (c) Decompose hash encoder, (d) Decompose geometry net with hard constraint.}
    \label{fig:different_arch}
    \vspace{-14pt}
\end{figure}

\mypara{Analysis of the misalignment from the network design.}
During the early stages of exploring the multimodal problem, we conducted various experiments on the network architecture. We attempted to enhance the network's capacity by increasing the hash encoding resolution and feature dimensions, as well as adjusting the depth and width of the MLP network
and increasing training iterations
to handle multimodal inputs.
However, these attempts did not yield significant improvements, as the underlying issue was not identified.
Subsequently, we delved into the network structure design, as illustrated in \cref{fig:different_arch} and the results are present on \cref{tab:different_arch}.
In \cref{fig:different_arch} (a), we decomposed the geometry MLP and surprisingly found that separating the geometry MLP resulted in almost no interference between the two modalities. However, this separation also meant that they did not mutually enhance each other. To further investigate the issue with the geometry net, in \cref{fig:different_arch} (b), we designed a shared geometry MLP but allowed the network to separately output two density values for each modality. We observed that this approach also enabled independent learning for both modalities, leading us to identify the issue as the misalignment in the density values, or more specifically, geometry misalignment.\begin{table}[t]
  \centering \small
  \caption{\textbf{Analysis of misalignment from the network design.}}
  \addtolength{\tabcolsep}{-1.9pt}
    \begin{tabularx}{\linewidth}{l|cc|cc}
  \toprule
 \multirow{2}{*}{Method}   & \multicolumn{2}{c|}{RGB Metric} & \multicolumn{2}{c}{LiDAR Metric}  \\
      \cmidrule(r){2-5} & PSNR$\uparrow$ & SSIM$\uparrow$  & C-D$\downarrow$ & F-score$\uparrow$ \\
      \midrule
     Single Modality & 24.45 &0.787 &0.088 & 0.920\\
     \midrule
      Decompose Geometry-net & 24.43& 0.784 & 0.084 & 0.929  \\
      Decompose Densities & 24.56& 0.788 & 0.084 & 0.929 \\
     \texttt{+} Hard Constraint & 24.85& 0.806 &0.089 &0.926\\ 
     Decompose Hash-encoder& 24.42 &0.793 &0.087 & 0.931\\
      \texttt{+} Hard Constraint & 24.53& 0.793 &0.084 & 0.929\\
      \bottomrule
    \end{tabularx}
  \label{tab:different_arch}
  \vspace{-10pt}
\end{table}
Having identified the problem as the geometry misalignment, we proceeded to decompose the source hash decoding as \cref{fig:different_arch} (c), and once again observed that the two modalities did not affect each other.
Furthermore, building upon the (b) and (c), we also tried to incorporate an alignment technique similar to the NeRF distillation work~\cite{fang2023nerfkd}. Specifically, we directly aligned the density values of the two separate networks as \cref{fig:different_arch} (d). Although this hard constraint demonstrated some improvement in the outcomes, it remained a forced alignment and failed to address the underlying conflict. Consequently, the performance improvement was limited as in \cref{tab:different_arch}.
Through these efforts in model network structure design, we not only verified the geometry misalignment issue but also inspired the design of our \name{}, which involves decomposing hash encoding and aligning coarse geometry.

\mypara{Ablations on the proposed modules of \name{}.}
In \cref{tab:ablations_all}, we conduct an ablation study to analyze the effects of the proposed components. Specifically, we observe that both the \Geofusion{} and \Geoinit{} modules improve the multimodal performance, which aligns with the qualitative analysis presented earlier in \cref{fig:geo_fusion} and \cref{fig:geo_init}. Moreover, combining the two modules leads to further improvements, indicating that they effectively alleviate the misalignment issue.

\begin{table}[t]
  \centering \small
  \caption{\textbf{Ablation study for the proposed components.}
  }
  \addtolength{\tabcolsep}{-2.4pt}
    \begin{tabularx}{\linewidth}{l|cc|cc}
  \toprule
 \multirow{2}{*}{Method}   & \multicolumn{2}{c|}{RGB Metric} & \multicolumn{2}{c}{LiDAR Metric}  \\
      \cmidrule(r){2-5}   & PSNR$\uparrow$ & SSIM$\uparrow$ & C-D$\downarrow$ & F-score$\uparrow$ \\
      \midrule

        Single Modality & 24.45 & 0.787 &0.088 & 0.920\\

     \midrule
        \multicolumn{3}{l}{\textit{\textbf{\Geoinit}}}    \\
        \midrule

     Load LiDAR Encoder & 24.34& 0.784 & 0.096 & 0.920  \\
     \texttt{+} Fix LiDAR Encoder  & 18.60 & 0.537 & 0.120 & 0.904  \\
     Detach RGB Density & 12.41 & 0.597 & 0.098 & 0.916  \\

     \midrule
        \multicolumn{3}{l}{\textit{\textbf{\Geofusion}}}    \\
        \midrule

        Addition& 24.47 &0.808 & 0.081& 0.929 \\
        Attention& 24.48&  0.791&  0.089 &0.929\\
        Concatenation & 24.64 & 0.810 & 0.079&0.930\\
        Share Coarse-Geo& 24.45 &0.794  & 0.087 &0.930\\

    \midrule
        \multicolumn{3}{l}{\textit{\textbf{\name{}}}}    \\
        \midrule

      w/  \Geoinit &24.70 \plus &0.806 &0.079 \plus &0.930  \\
        w/  \Geofusion &24.64 \plus& 0.810 & 0.079 \plus&0.930 \\
        \name{}   &\textbf{25.20 \plus} &\textbf{0.816}& \textbf{0.077 \plus}&\textbf{0.932}  \\

      \bottomrule
    \end{tabularx}
  \label{tab:ablations_all}
   \vspace{-10pt}
\end{table}

We also provide an investigation of the design of these modules.
For \Geoinit{}, we explored different initialization strategies. One straightforward approach was to directly load a pre-trained LiDAR encoder as the initialization for geometry, referred to as \textit{Load LiDAR Encoder}. However, this approach did not yield better results due to misalignment issues. Then we considered fixing the pre-trained geometry, i.e., \textit{Fix LiDAR Encoder}, to mitigate the interference caused by noisy camera geometry, and only the MLPs were trained, similar to the training process in CLONeR~\cite{carlson2023cloner}. Nevertheless, we got unfavorable results as the two modalities had different FOVs, resulting in the training being effective only in the pre-trained LiDAR FOV as shown in \cref{fig:fov} of the appendix. Moreover, the analyzed misalignment depicted in \cref{fig:migalign_sensor} also contributes to obstacles.
Additionally, we also attempted to detach the gradient of density from the camera modality to avoid geometry conflicts, which is similar to gradients blocking in Panoptic-Lifting~\cite{siddiqui2023panoptic}, but the FOV mismatch and misalignment issue persisted. Hence, none of these designs proved as effective as our proposed \Geoinit{}, which provides shared initial coarse geometry while allowing the learning of hash features from both modalities to capture their respective details.

For \Geofusion{}, we explored different fusion strategies for alignment, including addition, concatenation, and attention mechanisms. We adopted the efficient attention structure from ER-NeRF~\cite{li2023efficient_attn}.
Improvements can be observed with each fusion approach, and we ultimately selected concatenation
as it's the most effective.
Moreover, further exploration of more powerful fusion modules for alignment remains a promising research direction. Additionally, it's also considered to share the coarse geometry in GAA rather than aligning, however, the results obtained are not satisfactory. As also observed by Panoptic-Lifting~\cite{siddiqui2023panoptic}, despite the underlying scene geometry being the same, features required for different modalities and representations might be slightly different, e.g., the LiDAR intensity and image color.
Thus, similar to other works~\cite{smitt2023pag, rosu2023permutosdf}, we employ a separate grid encoder for each specific modality.

\begin{table}[t]
  \centering \small
  \caption{\textbf{Ablation study for the coarse geometry levels.}
  }
  \addtolength{\tabcolsep}{-0.1pt}
    \begin{tabularx}{\linewidth}{l|c|cc|cc}
  \toprule
 \multirow{2}{*}{Dataset} &
 \multirow{2}{*}{Levels}  & \multicolumn{2}{c|}{RGB Metric} & \multicolumn{2}{c}{LiDAR Metric}  \\
      \cmidrule(r){3-6} &    & PSNR$\uparrow$ & SSIM$\uparrow$  & C-D$\downarrow$ & F-score$\uparrow$  \\
      \midrule
     &4 & 23.68& 0.773 &  0.080&0.929\\
       \multirow{1}{*}{KITTI-360}  &\underline{ 8 }  & \underline{25.20}& \underline{0.816} & \underline{0.077}& \underline{0.932} \\
         & 16&  24.49& 0.803  & 0.091 & 0.921 \\
\midrule
     &6  & 28.73 &0.834  & 0.156 &0.892\\
       \multirow{1}{*}{Waymo}  &\underline{ 9 }&\underline{29.22} & \underline{0.841} & \underline{0.151}& \underline{0.896} \\
         & 12& 28.72 & 0.835 & 0.155 & 0.891\\
      \bottomrule

    \end{tabularx}
  \label{tab:ablations_levels}
  \vspace{-10pt}
\end{table}

\mypara{Ablations on the coarse geometry levels.}
As shown in \cref{tab:ablations_levels}, for each dataset, we search for the optimal alignment level, i.e., the $\beta$, of the coarse geometry.
As indicated by the dynamic network technology~\cite{han2021dynamic}, developing dynamic search levels holds the potential for further improvements in the alignment process, and we left it as future work.

\section{Conclusion}
In this paper, we thoroughly investigated and validated the
misalignment issue in multimodal NeRF through various analyses, such as the examination and visualization of raw sensor inputs, hash features, and density values, as well as experiments on various network architectures.
Futhermore, we propose \name{}, with two simple yet effective modules, Geometry-Aware Alignment (\Geofusion{}) and Shared Geometry Initialization (\Geoinit{}), to address the misalignment issue by aligning the consistent coarse geometry of different modalities while preserving their unique details.
We conduct extensive experiments on multiple datasets and scenes and demonstrate the effectiveness of our proposed method in improving multimodal fusion and alignment within a unified NeRF framework.
We hope that our work can inspire future research in the field of multimodal NeRF.



\mypara{Acknowledgements}. 
This work was supported in part by the National Key R\&D Program of China under Grant No. 2020AAA0109700,  Guangdong Outstanding Youth Fund (Grant No. 2021B1515020061), Mobility Grant Award under Grant No.  M-0461,  Shenzhen Science and Technology Program (Grant No. RCYX20200714114642083),  Shenzhen Science and Technology Program (Grant No. GJHZ20220913142600001), Nansha Key RD Program under Grant No.2022ZD014. 

{
    \small
    \bibliographystyle{ieeenat_fullname}
    \bibliography{main}

\begin{thebibliography}{64}
\providecommand{\natexlab}[1]{#1}
\providecommand{\url}[1]{\texttt{#1}}
\expandafter\ifx\csname urlstyle\endcsname\relax
  \providecommand{\doi}[1]{doi: #1}\else
  \providecommand{\doi}{doi: \begingroup \urlstyle{rm}\Url}\fi

\bibitem[Bai et~al.(2022)Bai, Hu, Zhu, Huang, Chen, Fu, and Tai]{bai2022transfusion}
Xuyang Bai, Zeyu Hu, Xinge Zhu, Qingqiu Huang, Yilun Chen, Hongbo Fu, and Chiew-Lan Tai.
\newblock Transfusion: Robust lidar-camera fusion for 3d object detection with transformers.
\newblock In \emph{Proceedings of the IEEE/CVF conference on computer vision and pattern recognition}, pages 1090--1099, 2022.

\bibitem[Barron et~al.(2021)Barron, Mildenhall, Tancik, Hedman, Martin-Brualla, and Srinivasan]{barron2021mip}
Jonathan~T Barron, Ben Mildenhall, Matthew Tancik, Peter Hedman, Ricardo Martin-Brualla, and Pratul~P Srinivasan.
\newblock Mip-nerf: A multiscale representation for anti-aliasing neural radiance fields.
\newblock In \emph{ICCV}, 2021.

\bibitem[Barron et~al.(2023)Barron, Mildenhall, Verbin, Srinivasan, and Hedman]{barron2023zip}
Jonathan~T Barron, Ben Mildenhall, Dor Verbin, Pratul~P Srinivasan, and Peter Hedman.
\newblock Zip-nerf: Anti-aliased grid-based neural radiance fields.
\newblock \emph{ICCV}, 2023.

\bibitem[Carlson et~al.(2023)Carlson, Ramanagopal, Tseng, Johnson-Roberson, Vasudevan, and Skinner]{carlson2023cloner}
Alexandra Carlson, Manikandasriram~S Ramanagopal, Nathan Tseng, Matthew Johnson-Roberson, Ram Vasudevan, and Katherine~A Skinner.
\newblock Cloner: Camera-lidar fusion for occupancy grid-aided neural representations.
\newblock \emph{IEEE Robotics and Automation Letters}, 2023.

\bibitem[Chen et~al.(2022)Chen, Xu, Geiger, Yu, and Su]{chen2022tensorf}
Anpei Chen, Zexiang Xu, Andreas Geiger, Jingyi Yu, and Hao Su.
\newblock Tensorf: Tensorial radiance fields.
\newblock In \emph{ECCV}, 2022.

\bibitem[Datta et~al.(2023)Datta, Marshall, Dong, Li, and Nowrouzezahrai]{datta2023efficient}
Sayantan Datta, Carl Marshall, Zhao Dong, Zhengqin Li, and Derek Nowrouzezahrai.
\newblock Efficient graphics representation with differentiable indirection.
\newblock \emph{arXiv:2309.08387}, 2023.

\bibitem[Deng et~al.(2022)Deng, Liu, Zhu, and Ramanan]{deng2022dsnerf}
Kangle Deng, Andrew Liu, Jun-Yan Zhu, and Deva Ramanan.
\newblock Depth-supervised nerf: Fewer views and faster training for free.
\newblock In \emph{CVPR}, 2022.

\bibitem[Dosovitskiy et~al.(2017)Dosovitskiy, Ros, Codevilla, Lopez, and Koltun]{dosovitskiy2017carla}
Alexey Dosovitskiy, German Ros, Felipe Codevilla, Antonio Lopez, and Vladlen Koltun.
\newblock Carla: An open urban driving simulator.
\newblock In \emph{Conference on robot learning}, 2017.

\bibitem[Fang et~al.(2023)Fang, Xu, Wang, Yang, Wang, and Zhou]{fang2023nerfkd}
Shuangkang Fang, Weixin Xu, Heng Wang, Yi Yang, Yufeng Wang, and Shuchang Zhou.
\newblock One is all: Bridging the gap between neural radiance fields architectures with progressive volume distillation.
\newblock In \emph{AAAI}, 2023.

\bibitem[Fu et~al.(2022)Fu, Zhang, Chen, Lu, Zhu, Zhou, Geiger, and Liao]{fu2022panoptic}
Xiao Fu, Shangzhan Zhang, Tianrun Chen, Yichong Lu, Lanyun Zhu, Xiaowei Zhou, Andreas Geiger, and Yiyi Liao.
\newblock Panoptic nerf: 3d-to-2d label transfer for panoptic urban scene segmentation.
\newblock \emph{arXiv:2203.15224}, 2022.

\bibitem[Guo et~al.(2023)Guo, Deng, Li, Bai, Shi, Wang, Ding, Wang, and Li]{guo2023streetsurf}
Jianfei Guo, Nianchen Deng, Xinyang Li, Yeqi Bai, Botian Shi, Chiyu Wang, Chenjing Ding, Dongliang Wang, and Yikang Li.
\newblock Streetsurf: Extending multi-view implicit surface reconstruction to street views.
\newblock \emph{arXiv:2306.04988}, 2023.

\bibitem[Han et~al.(2021)Han, Huang, Song, Yang, Wang, and Wang]{han2021dynamic}
Yizeng Han, Gao Huang, Shiji Song, Le Yang, Honghui Wang, and Yulin Wang.
\newblock Dynamic neural networks: A survey.
\newblock \emph{TPAMI}, 2021.

\bibitem[Hu et~al.(2023{\natexlab{a}})Hu, Wang, Ma, Yang, Gao, Liu, and Ma]{hu2023tri}
Wenbo Hu, Yuling Wang, Lin Ma, Bangbang Yang, Lin Gao, Xiao Liu, and Yuewen Ma.
\newblock Tri-miprf: Tri-mip representation for efficient anti-aliasing neural radiance fields.
\newblock In \emph{ICCV}, 2023{\natexlab{a}}.

\bibitem[Hu et~al.(2023{\natexlab{b}})Hu, Xiong, Zang, Jia, Han, and Ma]{hu2023pcnerf}
Xiuzhong Hu, Guangming Xiong, Zheng Zang, Peng Jia, Yuxuan Han, and Junyi Ma.
\newblock Pc-nerf: Parent-child neural radiance fields under partial sensor data loss in autonomous driving environments.
\newblock \emph{arXiv:2310.00874}, 2023{\natexlab{b}}.

\bibitem[Huang et~al.(2023{\natexlab{a}})Huang, Gojcic, Wang, Williams, Kasten, Fidler, Schindler, and Litany]{huang2023neural}
Shengyu Huang, Zan Gojcic, Zian Wang, Francis Williams, Yoni Kasten, Sanja Fidler, Konrad Schindler, and Or Litany.
\newblock Neural lidar fields for novel view synthesis.
\newblock \emph{ICCV}, 2023{\natexlab{a}}.

\bibitem[Huang et~al.(2023{\natexlab{b}})Huang, Zhang, Feng, Li, Wang, and Wang]{huang2023local}
Xin Huang, Qi Zhang, Ying Feng, Xiaoyu Li, Xuan Wang, and Qing Wang.
\newblock Local implicit ray function for generalizable radiance field representation.
\newblock In \emph{CVPR}, 2023{\natexlab{b}}.

\bibitem[Kingma and Ba(2014)]{kingma2014adam}
Diederik~P Kingma and Jimmy Ba.
\newblock Adam: A method for stochastic optimization.
\newblock \emph{arXiv:1412.6980}, 2014.

\bibitem[Lao et~al.(2023)Lao, Xu, Cai, Liu, and Zhao]{lao2023corresnerf}
Yixing Lao, Xiaogang Xu, Zhipeng Cai, Xihui Liu, and Hengshuang Zhao.
\newblock Corresnerf: Image correspondence priors for neural radiance fields.
\newblock In \emph{NeurIPS}, 2023.

\bibitem[Li et~al.(2023{\natexlab{a}})Li, Zhang, Bai, Zhou, and Gu]{li2023efficient_attn}
Jiahe Li, Jiawei Zhang, Xiao Bai, Jun Zhou, and Lin Gu.
\newblock Efficient region-aware neural radiance fields for high-fidelity talking portrait synthesis.
\newblock In \emph{ICCV}, 2023{\natexlab{a}}.

\bibitem[Li et~al.(2023{\natexlab{b}})Li, Li, and Zhu]{li2023read}
Zhuopeng Li, Lu Li, and Jianke Zhu.
\newblock Read: Large-scale neural scene rendering for autonomous driving.
\newblock In \emph{AAAI}, 2023{\natexlab{b}}.

\bibitem[Li et~al.(2023{\natexlab{c}})Li, M{\"u}ller, Evans, Taylor, Unberath, Liu, and Lin]{li2023neuralangelo}
Zhaoshuo Li, Thomas M{\"u}ller, Alex Evans, Russell~H Taylor, Mathias Unberath, Ming-Yu Liu, and Chen-Hsuan Lin.
\newblock Neuralangelo: High-fidelity neural surface reconstruction.
\newblock In \emph{CVPR}, 2023{\natexlab{c}}.

\bibitem[Liang et~al.(2023)Liang, Deng, Zhang, and Wang]{liang2023hr}
Erich Liang, Kenan Deng, Xi Zhang, and Chun-Kai Wang.
\newblock Hr-neus: Recovering high-frequency surface geometry via neural implicit surfaces.
\newblock \emph{arXiv:2302.06793}, 2023.

\bibitem[Liao et~al.(2022)Liao, Xie, and Geiger]{liao2022kitti360}
Yiyi Liao, Jun Xie, and Andreas Geiger.
\newblock Kitti-360: A novel dataset and benchmarks for urban scene understanding in 2d and 3d.
\newblock \emph{TPAMI}, 2022.

\bibitem[Lu et~al.(2023)Lu, Xu, Chen, Li, Lin, and Jiang]{lu2023dnmp}
Fan Lu, Yan Xu, Guang Chen, Hongsheng Li, Kwan-Yee Lin, and Changjun Jiang.
\newblock Urban radiance field representation with deformable neural mesh primitives.
\newblock In \emph{ICCV}, 2023.

\bibitem[Meuleman et~al.(2023)Meuleman, Liu, Gao, Huang, Kim, Kim, and Kopf]{meuleman2023progressively}
Andreas Meuleman, Yu-Lun Liu, Chen Gao, Jia-Bin Huang, Changil Kim, Min~H Kim, and Johannes Kopf.
\newblock Progressively optimized local radiance fields for robust view synthesis.
\newblock In \emph{CVPR}, 2023.

\bibitem[Mildenhall et~al.(2021)Mildenhall, Srinivasan, Tancik, Barron, Ramamoorthi, and Ng]{mildenhall2021nerf}
Ben Mildenhall, Pratul~P Srinivasan, Matthew Tancik, Jonathan~T Barron, Ravi Ramamoorthi, and Ren Ng.
\newblock Nerf: Representing scenes as neural radiance fields for view synthesis.
\newblock \emph{Communications of the ACM}, 2021.

\bibitem[M{\"u}ller et~al.(2022)M{\"u}ller, Evans, Schied, and Keller]{muller2022instantNGP}
Thomas M{\"u}ller, Alex Evans, Christoph Schied, and Alexander Keller.
\newblock Instant neural graphics primitives with a multiresolution hash encoding.
\newblock \emph{ToG}, 2022.

\bibitem[Neff et~al.(2021)Neff, Stadlbauer, Parger, Kurz, Mueller, Chaitanya, Kaplanyan, and Steinberger]{neff2021donerf}
Thomas Neff, Pascal Stadlbauer, Mathias Parger, Andreas Kurz, Joerg~H Mueller, Chakravarty R~Alla Chaitanya, Anton Kaplanyan, and Markus Steinberger.
\newblock Donerf: Towards real-time rendering of compact neural radiance fields using depth oracle networks.
\newblock In \emph{CGF}, 2021.

\bibitem[Ost et~al.(2021)Ost, Mannan, Thuerey, Knodt, and Heide]{ost2021NSG}
Julian Ost, Fahim Mannan, Nils Thuerey, Julian Knodt, and Felix Heide.
\newblock Neural scene graphs for dynamic scenes.
\newblock In \emph{CVPR}, 2021.

\bibitem[Peng et~al.(2022)Peng, Wei, Deng, Wang, and Hu]{peng2022balanced}
Xiaokang Peng, Yake Wei, Andong Deng, Dong Wang, and Di Hu.
\newblock Balanced multimodal learning via on-the-fly gradient modulation.
\newblock In \emph{CVPR}, 2022.

\bibitem[Rematas et~al.(2022)Rematas, Liu, Srinivasan, Barron, Tagliasacchi, Funkhouser, and Ferrari]{rematas2022urban-nerf}
Konstantinos Rematas, Andrew Liu, Pratul~P Srinivasan, Jonathan~T Barron, Andrea Tagliasacchi, Thomas Funkhouser, and Vittorio Ferrari.
\newblock Urban radiance fields.
\newblock In \emph{CVPR}, 2022.

\bibitem[Roessle et~al.(2022)Roessle, Barron, Mildenhall, Srinivasan, and Nie{\ss}ner]{roessle2022depthpriors}
Barbara Roessle, Jonathan~T Barron, Ben Mildenhall, Pratul~P Srinivasan, and Matthias Nie{\ss}ner.
\newblock Dense depth priors for neural radiance fields from sparse input views.
\newblock In \emph{CVPR}, 2022.

\bibitem[Rosu and Behnke(2023)]{rosu2023permutosdf}
Radu~Alexandru Rosu and Sven Behnke.
\newblock Permutosdf: Fast multi-view reconstruction with implicit surfaces using permutohedral lattices.
\newblock In \emph{CVPR}, 2023.

\bibitem[Sanders(2016)]{sanders2016UE4}
Andrew Sanders.
\newblock \emph{An introduction to Unreal engine 4}.
\newblock AK Peters/CRC Press, 2016.

\bibitem[Shah et~al.(2018)Shah, Dey, Lovett, and Kapoor]{shah2018airsim}
Shital Shah, Debadeepta Dey, Chris Lovett, and Ashish Kapoor.
\newblock Airsim: High-fidelity visual and physical simulation for autonomous vehicles.
\newblock In \emph{Field and Service Robotics: Results of the 11th International Conference}, pages 621--635. Springer, 2018.

\bibitem[Siddiqui et~al.(2023)Siddiqui, Porzi, Bul{\`o}, M{\"u}ller, Nie{\ss}ner, Dai, and Kontschieder]{siddiqui2023panoptic}
Yawar Siddiqui, Lorenzo Porzi, Samuel~Rota Bul{\`o}, Norman M{\"u}ller, Matthias Nie{\ss}ner, Angela Dai, and Peter Kontschieder.
\newblock Panoptic lifting for 3d scene understanding with neural fields.
\newblock In \emph{CVPR}, 2023.

\bibitem[Smitt et~al.(2023)Smitt, Halstead, Zimmer, L{\"a}be, Guclu, Stachniss, and McCool]{smitt2023pag}
Claus Smitt, Michael Halstead, Patrick Zimmer, Thomas L{\"a}be, Esra Guclu, Cyrill Stachniss, and Chris McCool.
\newblock Pag-nerf: Towards fast and efficient end-to-end panoptic 3d representations for agricultural robotics.
\newblock \emph{arXiv:2309.05339}, 2023.

\bibitem[Sun et~al.(2020)Sun, Kretzschmar, Dotiwalla, Chouard, Patnaik, Tsui, Guo, Zhou, Chai, Caine, et~al.]{sun2020waymo}
Pei Sun, Henrik Kretzschmar, Xerxes Dotiwalla, Aurelien Chouard, Vijaysai Patnaik, Paul Tsui, James Guo, Yin Zhou, Yuning Chai, Benjamin Caine, et~al.
\newblock Scalability in perception for autonomous driving: Waymo open dataset.
\newblock In \emph{CVPR}, 2020.

\bibitem[Sun et~al.(2021)Sun, Mai, and Hu]{sun2021learning}
Ya Sun, Sijie Mai, and Haifeng Hu.
\newblock Learning to balance the learning rates between various modalities via adaptive tracking factor.
\newblock \emph{SPL}, 2021.

\bibitem[Tancik et~al.(2022)Tancik, Casser, Yan, Pradhan, Mildenhall, Srinivasan, Barron, and Kretzschmar]{tancik2022blocknerf}
Matthew Tancik, Vincent Casser, Xinchen Yan, Sabeek Pradhan, Ben Mildenhall, Pratul~P Srinivasan, Jonathan~T Barron, and Henrik Kretzschmar.
\newblock Block-nerf: Scalable large scene neural view synthesis.
\newblock In \emph{CVPR}, 2022.

\bibitem[Tancik et~al.(2023)Tancik, Weber, Ng, Li, Yi, Kerr, Wang, Kristoffersen, Austin, Salahi, Ahuja, McAllister, and Kanazawa]{nerfstudio}
Matthew Tancik, Ethan Weber, Evonne Ng, Ruilong Li, Brent Yi, Justin Kerr, Terrance Wang, Alexander Kristoffersen, Jake Austin, Kamyar Salahi, Abhik Ahuja, David McAllister, and Angjoo Kanazawa.
\newblock Nerfstudio: A modular framework for neural radiance field development.
\newblock In \emph{SIGGRAPH}, 2023.

\bibitem[Tao et~al.(2023)Tao, Gao, Wang, Chen, Hao, Liang, Salzmann, and Yu]{tao2023lidarnerf}
Tang Tao, Longfei Gao, Guangrun Wang, Peng Chen, Dayang Hao, Xiaodan Liang, Mathieu Salzmann, and Kaicheng Yu.
\newblock Lidar-nerf: Novel lidar view synthesis via neural radiance fields.
\newblock \emph{arXiv:2304.10406}, 2023.

\bibitem[Turki et~al.(2023)Turki, Zhang, Ferroni, and Ramanan]{turki2023suds}
Haithem Turki, Jason~Y Zhang, Francesco Ferroni, and Deva Ramanan.
\newblock Suds: Scalable urban dynamic scenes.
\newblock In \emph{CVPR}, 2023.

\bibitem[Wang et~al.(2023{\natexlab{a}})Wang, Sun, Liu, Wu, Shen, Wu, Dai, and Zhang]{wang2023digging}
Chen Wang, Jiadai Sun, Lina Liu, Chenming Wu, Zhelun Shen, Dayan Wu, Yuchao Dai, and Liangjun Zhang.
\newblock Digging into depth priors for outdoor neural radiance fields.
\newblock \emph{arXiv:2308.04413}, 2023{\natexlab{a}}.

\bibitem[Wang et~al.(2023{\natexlab{b}})Wang, Liu, Chen, Liu, Liu, Komura, Theobalt, and Wang]{wang2023f2}
Peng Wang, Yuan Liu, Zhaoxi Chen, Lingjie Liu, Ziwei Liu, Taku Komura, Christian Theobalt, and Wenping Wang.
\newblock F2-nerf: Fast neural radiance field training with free camera trajectories.
\newblock In \emph{CVPR}, 2023{\natexlab{b}}.

\bibitem[Wang et~al.(2020)Wang, Tran, and Feiszli]{wang2020makes}
Weiyao Wang, Du Tran, and Matt Feiszli.
\newblock What makes training multi-modal classification networks hard?
\newblock In \emph{CVPR}, 2020.

\bibitem[Wang et~al.(2004)Wang, Bovik, Sheikh, and Simoncelli]{wang2004ssim}
Zhou Wang, Alan~C Bovik, Hamid~R Sheikh, and Eero~P Simoncelli.
\newblock Image quality assessment: from error visibility to structural similarity.
\newblock \emph{TIP}, 2004.

\bibitem[Wei Jong~Yang(2023)]{cao2022Misalignment}
Guan Cheng~Lee Wei Jong~Yang.
\newblock Addressing data misalignment in image-lidar fusion on point cloud segmentation.
\newblock \emph{arXiv:2309.14932}, 2023.

\bibitem[Weng et~al.(2023)Weng, Man, Park, Yuan, O'Toole, and Kitani]{weng2023aiodrive}
Xinshuo Weng, Yunze Man, Jinhyung Park, Ye Yuan, Matthew O'Toole, and Kris~M Kitani.
\newblock All-in-one drive: A comprehensive perception dataset with high-density long-range point clouds.
\newblock 2023.

\bibitem[Wu et~al.(2023)Wu, Liu, Luo, Zhong, Chen, Xiao, Hou, Lou, Chen, Yang, et~al.]{wu2023mars}
Zirui Wu, Tianyu Liu, Liyi Luo, Zhide Zhong, Jianteng Chen, Hongmin Xiao, Chao Hou, Haozhe Lou, Yuantao Chen, Runyi Yang, et~al.
\newblock Mars: An instance-aware, modular and realistic simulator for autonomous driving.
\newblock \emph{arXiv:2307.15058}, 2023.

\bibitem[Xie et~al.(2023{\natexlab{a}})Xie, Gherardi, Pan, and Huang]{xie2023hollownerf}
Xiufeng Xie, Riccardo Gherardi, Zhihong Pan, and Stephen Huang.
\newblock Hollownerf: Pruning hashgrid-based nerfs with trainable collision mitigation.
\newblock In \emph{ICCV}, 2023{\natexlab{a}}.

\bibitem[Xie et~al.(2023{\natexlab{b}})Xie, Zhang, Li, Zhang, and Zhang]{xie2023s-nerf}
Ziyang Xie, Junge Zhang, Wenye Li, Feihu Zhang, and Li Zhang.
\newblock S-nerf: Neural radiance fields for street views.
\newblock \emph{arXiv:2303.00749}, 2023{\natexlab{b}}.

\bibitem[Yang et~al.(2023{\natexlab{a}})Yang, Ivanovic, Litany, Weng, Kim, Li, Che, Xu, Fidler, Pavone, et~al.]{yang2023emernerf}
Jiawei Yang, Boris Ivanovic, Or Litany, Xinshuo Weng, Seung~Wook Kim, Boyi Li, Tong Che, Danfei Xu, Sanja Fidler, Marco Pavone, et~al.
\newblock Emernerf: Emergent spatial-temporal scene decomposition via self-supervision.
\newblock \emph{arXiv:2311.02077}, 2023{\natexlab{a}}.

\bibitem[Yang et~al.(2023{\natexlab{b}})Yang, Yang, Guo, Xiong, Wang, and Liao]{yang2023urbangiraffe}
Yuanbo Yang, Yifei Yang, Hanlei Guo, Rong Xiong, Yue Wang, and Yiyi Liao.
\newblock Urbangiraffe: Representing urban scenes as compositional generative neural feature fields.
\newblock \emph{arXiv:2303.14167}, 2023{\natexlab{b}}.

\bibitem[Yang et~al.(2023{\natexlab{c}})Yang, Chen, Wang, Manivasagam, Ma, Yang, and Urtasun]{yang2023unisim}
Ze Yang, Yun Chen, Jingkang Wang, Sivabalan Manivasagam, Wei-Chiu Ma, Anqi~Joyce Yang, and Raquel Urtasun.
\newblock Unisim: A neural closed-loop sensor simulator.
\newblock In \emph{CVPR}, 2023{\natexlab{c}}.

\bibitem[Yu et~al.(2021{\natexlab{a}})Yu, Fridovich-Keil, Tancik, Chen, Recht, and Kanazawa]{yu2021plenoxels}
Alex Yu, Sara Fridovich-Keil, Matthew Tancik, Qinhong Chen, Benjamin Recht, and Angjoo Kanazawa.
\newblock Plenoxels: Radiance fields without neural networks.
\newblock \emph{arXiv:2112.05131}, 2021{\natexlab{a}}.

\bibitem[Yu et~al.(2021{\natexlab{b}})Yu, Ye, Tancik, and Kanazawa]{yu2021pixelnerf}
Alex Yu, Vickie Ye, Matthew Tancik, and Angjoo Kanazawa.
\newblock pixelnerf: Neural radiance fields from one or few images.
\newblock In \emph{CVPR}, 2021{\natexlab{b}}.

\bibitem[Yu et~al.(2022{\natexlab{a}})Yu, Tao, Xie, Lin, Wu, Xia, Liang, Sun, Deng, Hao, et~al.]{yu2022benchmarking}
Kaicheng Yu, Tang Tao, Hongwei Xie, Zhiwei Lin, Zhongwei Wu, Zhongyu Xia, Tingting Liang, Haiyang Sun, Jiong Deng, Dayang Hao, et~al.
\newblock Benchmarking the robustness of lidar-camera fusion for 3d object detection.
\newblock \emph{CVPRW}, 2022{\natexlab{a}}.

\bibitem[Yu et~al.(2022{\natexlab{b}})Yu, Peng, Niemeyer, Sattler, and Geiger]{yu2022monosdf}
Zehao Yu, Songyou Peng, Michael Niemeyer, Torsten Sattler, and Andreas Geiger.
\newblock Monosdf: Exploring monocular geometric cues for neural implicit surface reconstruction.
\newblock \emph{NeurIPS}, 2022{\natexlab{b}}.

\bibitem[Zhang et~al.(2023)Zhang, Zhang, Kuang, and Zhang]{zhang2023nerflidar}
Junge Zhang, Feihu Zhang, Shaochen Kuang, and Li Zhang.
\newblock Nerf-lidar: Generating realistic lidar point clouds with neural radiance fields.
\newblock \emph{ICCV}, 2023.

\bibitem[Zhang et~al.(2018)Zhang, Isola, Efros, Shechtman, and Wang]{zhang2018lpips}
Richard Zhang, Phillip Isola, Alexei~A Efros, Eli Shechtman, and Oliver Wang.
\newblock The unreasonable effectiveness of deep features as a perceptual metric.
\newblock In \emph{CVPR}, 2018.

\bibitem[Zheng et~al.(2023{\natexlab{a}})Zheng, Bao, Hebert, and Wang]{zheng2023multi}
Shuhong Zheng, Zhipeng Bao, Martial Hebert, and Yu-Xiong Wang.
\newblock Multi-task view synthesis with neural radiance fields.
\newblock In \emph{ICCV}, 2023{\natexlab{a}}.

\bibitem[Zheng et~al.(2023{\natexlab{b}})Zheng, Wu, Lu, Lu, Chen, and Jiang]{zheng2023neuralpci}
Zehan Zheng, Danni Wu, Ruisi Lu, Fan Lu, Guang Chen, and Changjun Jiang.
\newblock Neuralpci: Spatio-temporal neural field for 3d point cloud multi-frame non-linear interpolation.
\newblock In \emph{CVPR}, 2023{\natexlab{b}}.

\bibitem[Zhi et~al.(2021)Zhi, Laidlow, Leutenegger, and Davison]{zhi2021semanticnerf}
Shuaifeng Zhi, Tristan Laidlow, Stefan Leutenegger, and Andrew~J Davison.
\newblock In-place scene labelling and understanding with implicit scene representation.
\newblock In \emph{ICCV}, 2021.

\end{thebibliography}
}

\clearpage
\setcounter{page}{1}
\maketitlesupplementary
\section{Limitations and Future work}
As this paper is the first to reveal the misalignment issue in multimodal learning in NeRF, there remains room for improvement, which we would like to address in future work. 
As such, it is better suited to static scenes. Fortunately, there have been notable advancements in handling dynamic scenes~\cite{yang2023unisim, wu2023mars, ost2021NSG, yang2023emernerf}, and our decomposed encoding formulation can be seamlessly integrated with these advances.
Moreover, for each dataset, it is necessary to search for the optimal alignment level of the coarse geometry. As indicated by the dynamic network technology~\cite{han2021dynamic}, developing dynamic search levels holds the potential for further improvements in the alignment process.
Furthermore, exploring more powerful fusion modules for alignment represents a promising research direction.
Another impact of our coarse geometry alignment is the incorporation of multiple hash encoders, which introduce additional model parameters and computation. Nonetheless, we optimize per scene with two to three hours on a single NVIDIA GeForce RTX 3090 GPU, which is still much more cost-effective compared to traditional handcrafted game-engine-based virtual worlds~\cite{dosovitskiy2017carla, sanders2016UE4, shah2018airsim},
 and there also have been efforts ~\cite{datta2023efficient, xie2023hollownerf} in improving the efficiency of hash encoders. 
Altogether, we hope that our work will inspire other researchers to contribute to the development of multimodal NeRF.

\begin{figure}[h]
    \centering
    \includegraphics[width=0.9\linewidth]{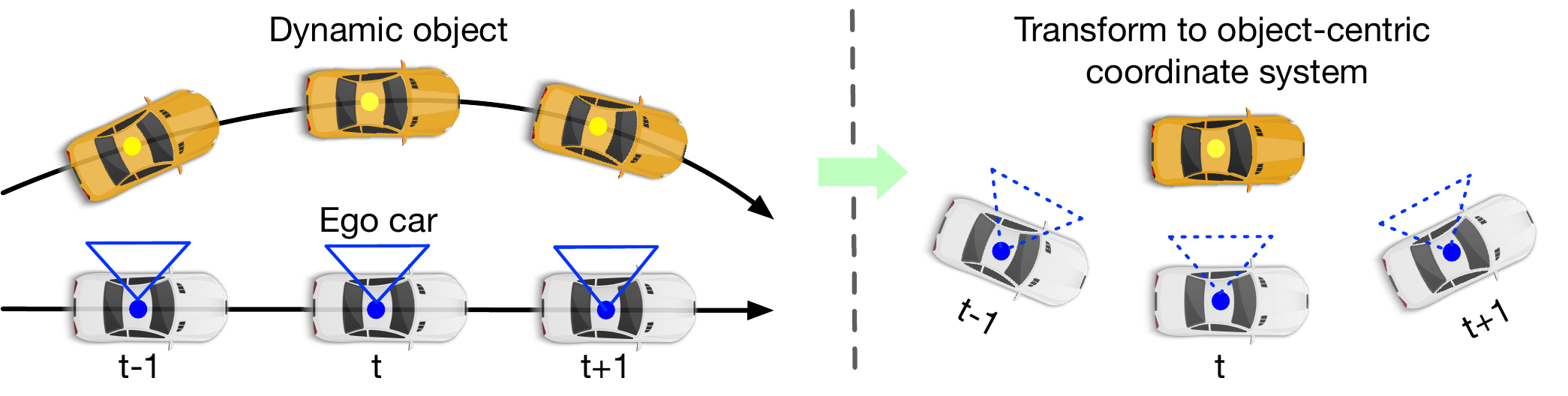}
    \vspace{-10pt}
    \caption{\textbf{The illustration of handling dynamic objects.} }
    \label{fig:dynamic}
\end{figure}

\mypara{Discuss dynamic foreground objects.} 
In \cref{fig:dynamic}, we illustrate the approach for dynamic scenes, where dynamic objects are modeled separately with the static background, and each object is transformed into its object-centroid coordinate system. This allows us to treat each object field as a small static scene, which can be directly extended to our \name.
Additionally, dynamic objects pose more challenges, and our \name{} can provide fusion models with different levels of alignment for the dynamic objects and static background, providing a comprehensive solution.

\section{Additional Details}
\label{sec:details}

\mypara{Dataset.} AIODrive consists of 100 video sequences generated by the CARLA Simulator, comprising about 100k labeled images and point cloud data. For our investigation, we utilize the provided mini-version of the dataset and masked the dynamic objects as Nerfstudio~\cite{nerfstudio}. We select every $15$th image in the sequences as the test set and take the remaining ones as the training set. 
KITTI-360 is a large-scale dataset containing over 320k images and 100k laser scans collected in urban environments with a driving distance of around $73.7$ km. We select 4 static suburb sequences as PNF~\cite{fu2022panoptic} and LiDAR-NeRF~\cite{tao2023lidarnerf}. Each sequence contains 64 frames, with 4 equidistant frames for evaluation.
For Waymo Open Dataset, we also select the 4 sequences mainly containing static objects for our experiments. We reserve every 10th frame as a test view and use the remaining about 188 samples for training.

\mypara{Implementation details.}
Our \name{} is implemented based on open-source LiDAR-NeRF~\cite{tao2023lidarnerf}. 
We optimize our \name{} model per scene with two to three hours of training time using a single NVIDIA GeForce RTX 3090 GPU.
We use Adam~\cite{kingma2014adam} with a learning rate of 1e-2 to train our models. The coarse and fine networks are sampled 768 and 64 samples per ray, respectively. The finest resolution of the hash encoding is set to 32768.
To better evaluate and compare the synthesis capability for details, 
we train and evaluate our methods and all the baselines with full-resolution images and LiDAR input. 
All ablation and analysis experiments were conducted on the sequence \textit{seq-1908-1971} of KITTI-360 dataset and the segment \textit{17761959194352517553\_5448\_420\_5468\_420} of  Waymo dataset, which both are large scenes with numerous objects, making them ideal sequences for comparison.

\begin{figure}[t]
    \centering
    \includegraphics[width= 1\linewidth]{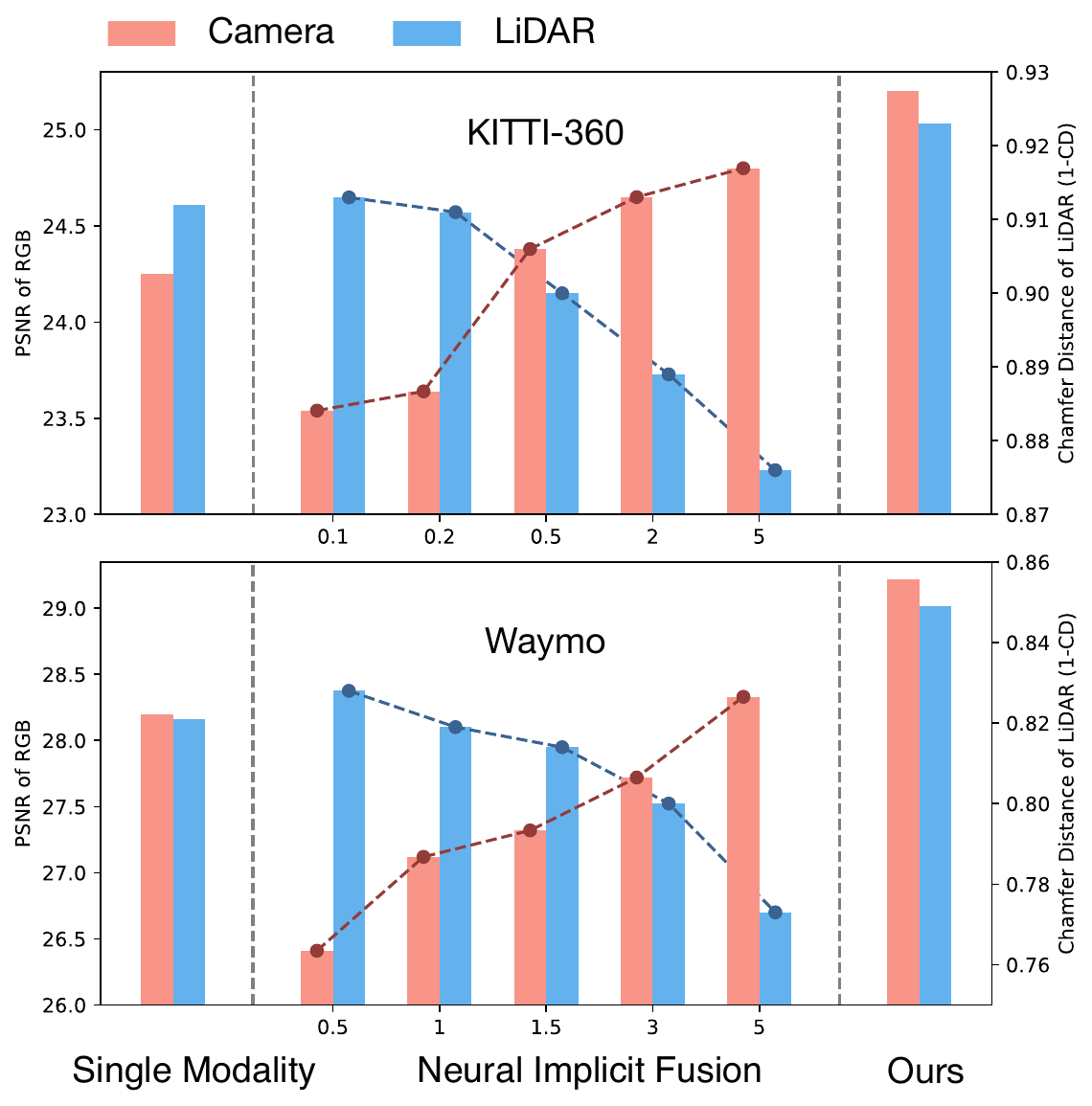}
    \caption{\textbf{The misalignment issue in multimodal implicit field.}
    For implicit neural fusion, there is a trade-off between the modalities due to the misalignment, making it challenging to improve both modalities simultaneously. Conversely, our method addresses the misalignment issue and achieves boosted multimodal performance.  The horizontal axis denotes the weight ratio between the camera and LiDAR modality, $\lambda_c / \lambda_l$.
    }
    \label{fig:teaser_old}
\end{figure}

\section{Additional Results}

\begin{figure}[t]
    \centering
    \includegraphics[width=1\linewidth]{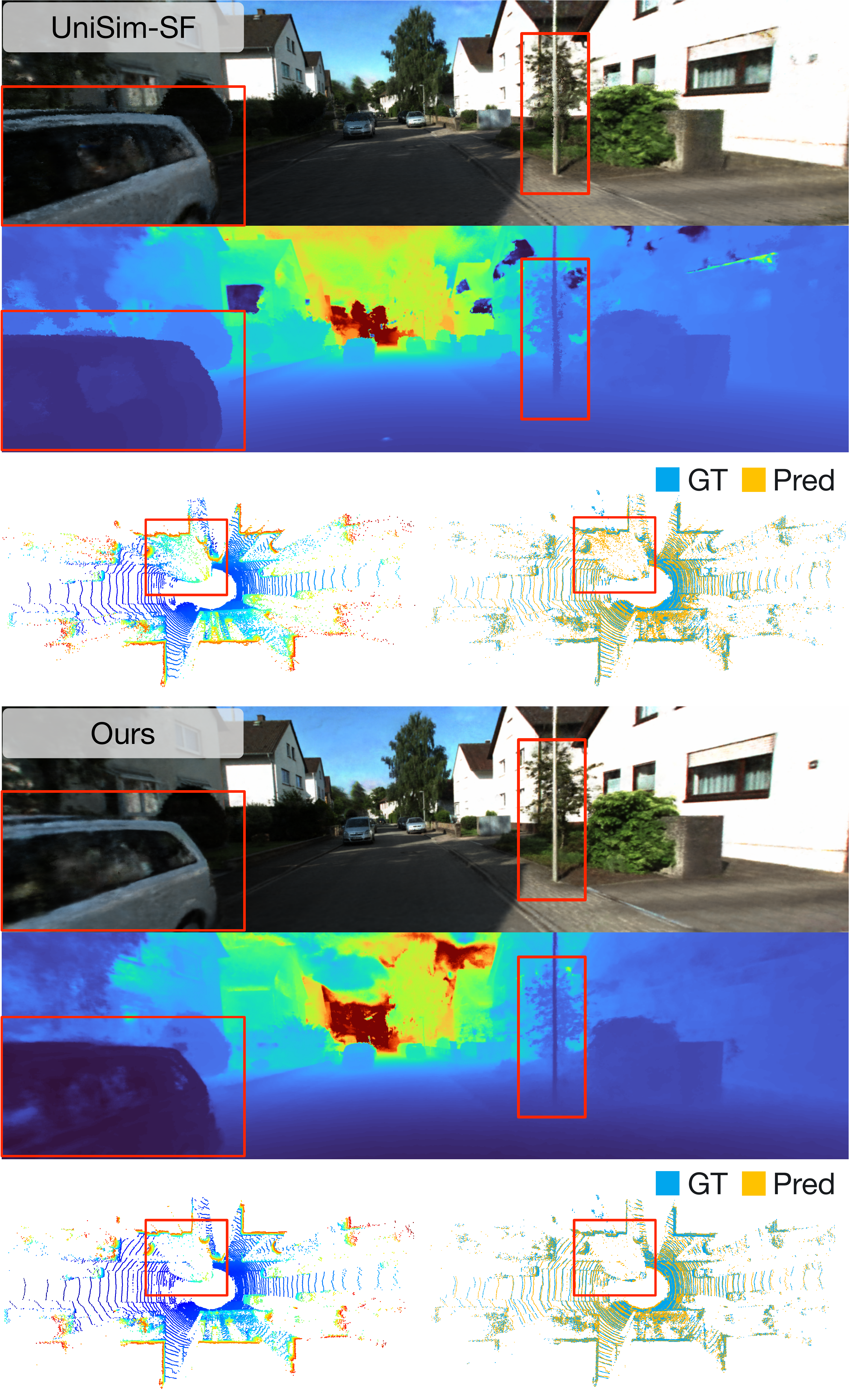}
    \caption{\textbf{The misalignment issue in multimodal implicit field.}  Implicit fusing misaligned modalities leads to suboptimal results, i.e., the blurred image and messy points, as depicted in the figure. Our \name{} enhances the alignment and fusion between LiDAR and camera modalities, leading to more accurate join synthesis of novel views (zoom-in for the best of views).}
    \label{fig:mif}
\end{figure}

\mypara{Misalignment issue in the multimodal implicit field.}
Previous multimodal implicit fields, e.g., UniSim~\cite{yang2023unisim}, explored fusing multiple modalities within a single field, aiming to share implicit features from different modalities to enhance performance. However, the misaligned modalities often contradict each other, and as shown in \cref{fig:teaser_old}, the line plot clearly illustrates a trade-off between the modalities: optimizing for one modality, such as the camera, can have a negative impact on the performance of another modality, such as LiDAR, and vice versa. 
In contrast, our \name{} effectively addresses the misalignment issue and achieves enhanced multimodal performance, as demonstrated by the bar chart in  \cref{fig:teaser_old}.
We also provide qualitative visualization in \cref{fig:mif}.

\begin{figure}[t]
    \centering
    \includegraphics[width=1\linewidth]{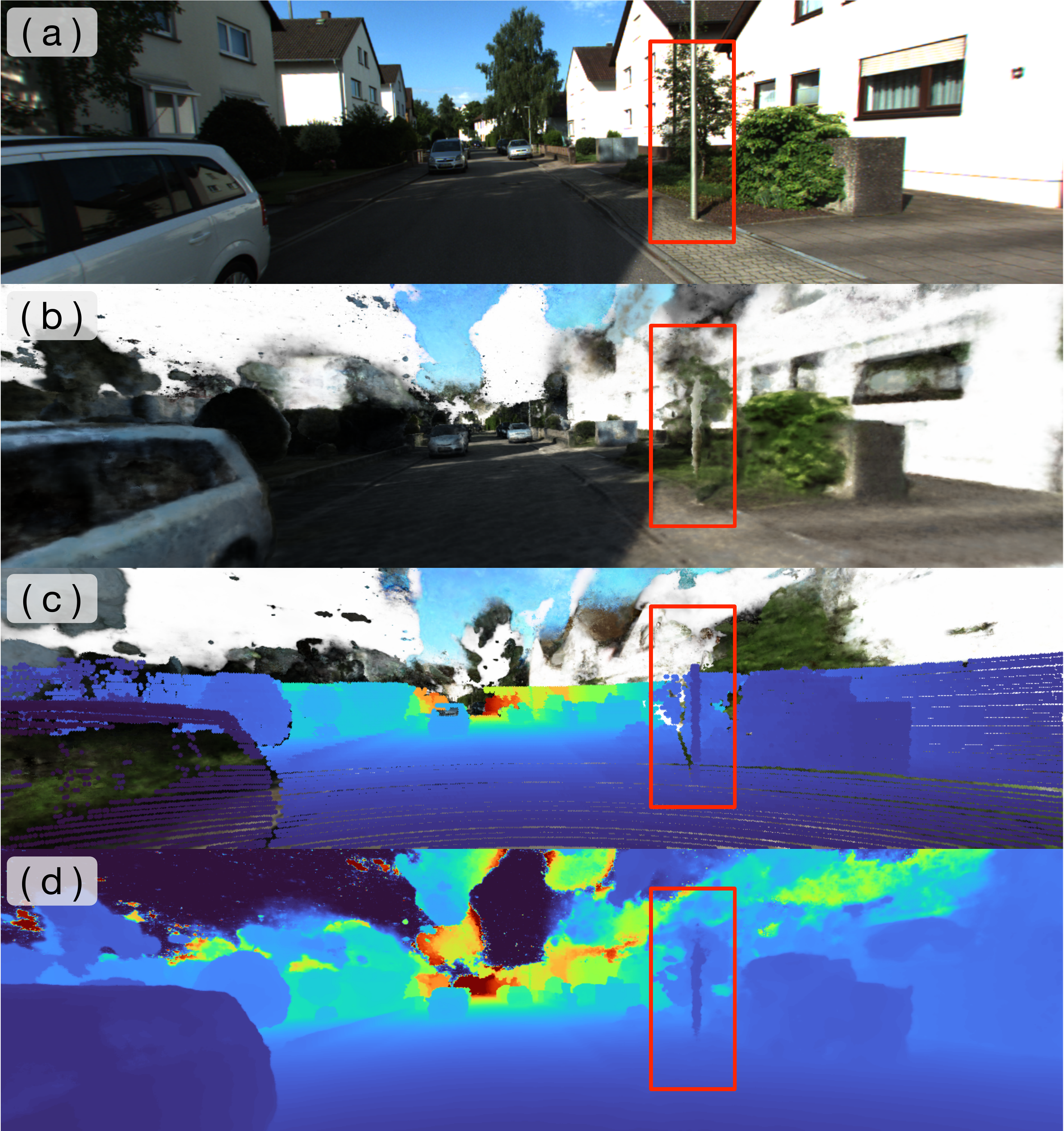}
    \caption{\textbf{Misalignment issue and different FOV of LiDAR and camera.} (a) Original image, (b) Rendered image, (c) Rendered image with projected points from associate LiDAR frame, (d) Rendered image depth.}
    \label{fig:fov}
\end{figure}

\mypara{Misalignment issue and different FOV.}
In \cref{subsec:ablation}, we attempted to detach the gradient of density from the camera modality to avoid geometry conflicts, i.e., \textit{Detach RGB Density}, which is similar to gradients blocking in Panoptic-Lifting~\cite{siddiqui2023panoptic}, but the FOV mismatch and misalignment issue have become significant obstacles as shown in \cref{fig:fov}.
From (a)(b)(c), we can observe that the two modalities had different FOVs, resulting in the training being effective only in the pre-trained LiDAR FOV. 
Moreover, as demonstrated in \cref{fig:migalign_sensor} of \cref{subsec:misalign} and (a)(c), LiDAR and camera exhibit variances in capturing finer details. Therefore, when solely relying on LiDAR for optimizing geometry, the resulting camera image and depth are both unsatisfactory, as illustrated by the distorted and thicker pole in (b) and (d). These figures further emphasize the importance of addressing the misalignment issue.

\begin{table}[h]
  \centering \small
  \caption{\textbf{State-of-the-art results on AIODrive dataset.}}
  \addtolength{\tabcolsep}{-3pt}
    \begin{tabularx}{\linewidth}{l|c|cc|cc}
  \toprule
 \multirow{2}{*}{Method} & \multirow{2}{*}{M}    & \multicolumn{2}{c|}{RGB Metric} & \multicolumn{2}{c}{LiDAR Metric}  \\
      \cmidrule(r){3-6} & & PSNR$\uparrow$ & SSIM$\uparrow$  & C-D$\downarrow$ & F-score$\uparrow$ \\
    \midrule
         i-NGP~\cite{muller2022instantNGP}& C & 34.43 &0.893 & --&--\\
     LiDAR-NeRF~\cite{tao2023lidarnerf}&L &--&--&0.178 & 0.873\\
     \midrule
      \UniSim ~\cite{yang2023unisim} & LC&  34.53 \plus  &  0.904 & 0.153 \plus &  0.905\\
        \name{} (Ours)& LC&\textbf{34.82 \plus}& 0.908&\textbf{ 0.123 \plus} & 0.915\\

      \bottomrule
      \multicolumn{6}{l}{\scriptsize{M, L, C denotes modality, LiDAR, camera respectively.}}
    \end{tabularx}
  \label{tab:aiodrive_sota}
\end{table}

\mypara{State-of-the-art results on AIODrive dataset.}
As shown in \cref{tab:aiodrive_sota}, our \name{}, achieves superior results even on the AIODrive synthetic dataset without the misalignment issue, outperforming the previous implicit fusion approach. 
This can be attributed to the fact that despite having the same underlying scene geometry, the required features for different modalities and representations might differ slightly, such as the LiDAR intensity and image color.
This also aligns with the results of \textit{Share Coarse-Geo} in \cref{tab:ablations_all} of \cref{subsec:ablation} and the observation in Panoptic-Lifting~\cite{siddiqui2023panoptic}.
These results both demonstrate the efficiency of our network design.

\mypara{Details results on KITTI-360 and Waymo datasets.}
We report detailed results on the sequences of the KITTI-360 and Waymo datasets in \cref{tab:all_res_full}. Our \name{} consistently outperforms the baselines over all sequences in all metrics. The details of sequences are also shown in \cref{tab:all_res_full}.

\mypara{Boosting downstream applications.}
We are eager to explore the potential benefits of improving downstream applications. We choose the powerful LiDAR-camera fusion detection method TransFusion~\cite{bai2022transfusion} and employ our AlignMiF to generate more diverse sensor data for data augmentation. Due to computational constraints, we conducted experiments on a limited number of Waymo scenes. The results in \cref{tab:detection} demonstrate the effectiveness of our approach in enhancing the performance of the downstream model.

\mypara{Computational complexities.} 
In \cref{tab:complexities}, we present the computational cost to facilitate further research.

\begin{table}[t]
  \centering \footnotesize
  \caption{\textbf{Enhancing the detection model with \name{}.}}
  \addtolength{\tabcolsep}{-4.9pt}
    \begin{tabularx}{0.97\linewidth}{l|c|c|c|c}
  \toprule
  Method & L1\_mAP & 	L1\_mAPH &	L2\_mAP	 & L2\_mAPH   \\
\midrule
  TransFusion~\cite{bai2022transfusion} & 38.71 &35.05 &33.98 & 30.79\\
   + \name{} & 40.18 \textcolor{cvprblueplusplus}{\scalebox{0.92}[0.92]{(\textbf{+1.47})}} & 36.39\textcolor{cvprblueplusplus}{\scalebox{0.92}[0.92]{(\textbf{+1.34})}}	& 35.31\textcolor{cvprblueplusplus}{\scalebox{0.92}[0.92]{(\textbf{+1.33})}} &32.01 \textcolor{cvprblueplusplus}{\scalebox{0.92}[0.92]{(\textbf{+1.22})}}\\
      \bottomrule
    \end{tabularx}
  \label{tab:detection}
\end{table}


\begin{table}[t]
  \centering \footnotesize
  \caption{\textbf{Computational cost (microsecond) for rendering 4096 rays.}}
  \addtolength{\tabcolsep}{-2.1pt}
    \begin{tabularx}{0.94\linewidth}{l|c |c | c}
  \toprule
  Method & Hash-Encoding &  Geo-MLP &  Color-MLP  \\
\midrule
 \UniSim ~\cite{yang2023unisim}& 86 & 72 & 96\\
 \name{} &{86 (SGI) + 156 (GAA)} & 72  & 96 \\

      \bottomrule
    \end{tabularx}
  \label{tab:complexities}
\end{table}

\subsection{Qualitative Results}

\begin{figure*}[h]
    \centering
    \includegraphics[width=1\linewidth]{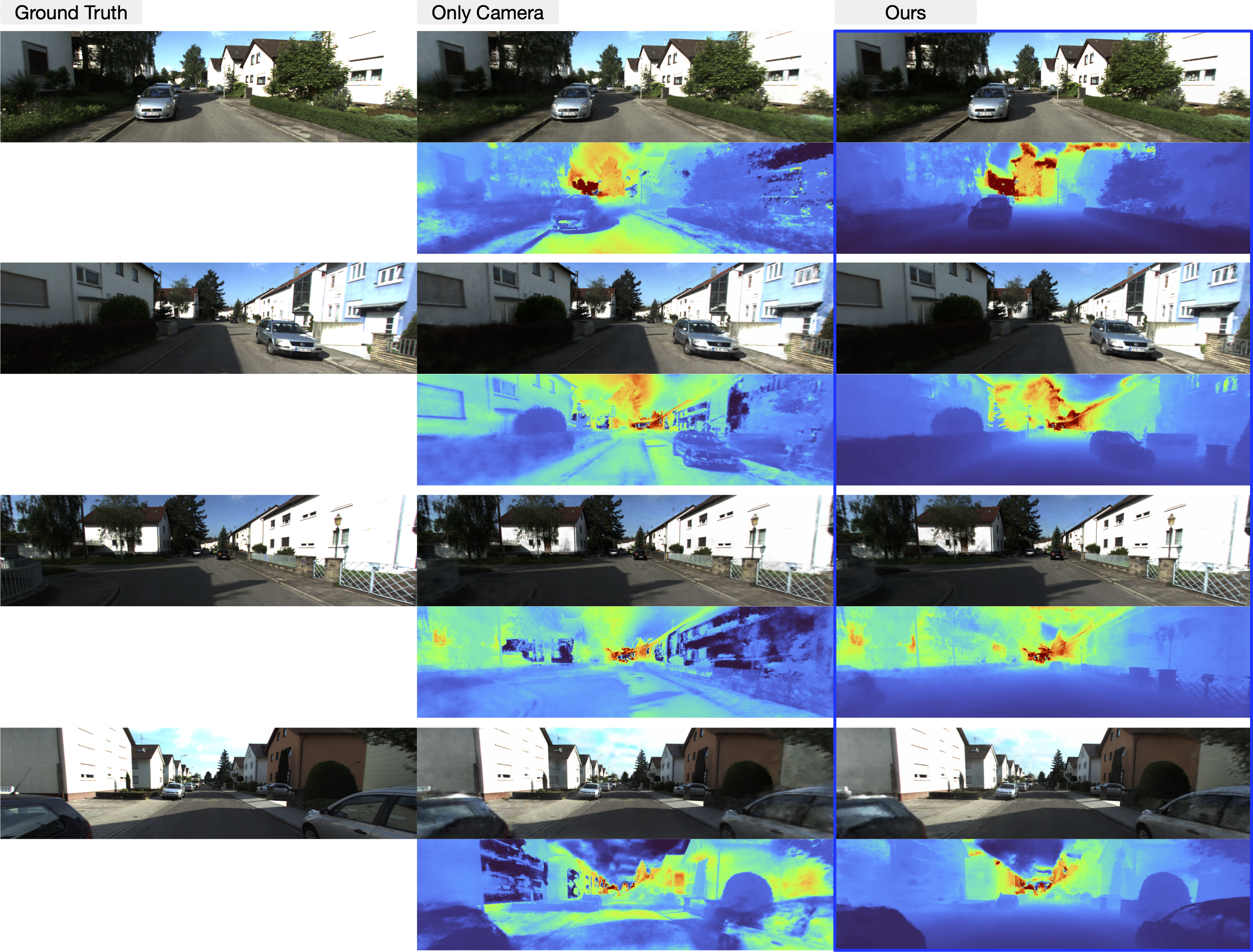}
    \caption{\textbf{Qualitative results of the camera on KITTI-360 dataset. } Our \name{} enhances information interactions between modalities and improves image and depth quality in the camera using LiDAR information.
    }
    \label{fig:res_kitti}
\end{figure*}

\begin{figure*}[h]
    \centering
    \includegraphics[width=1\linewidth]{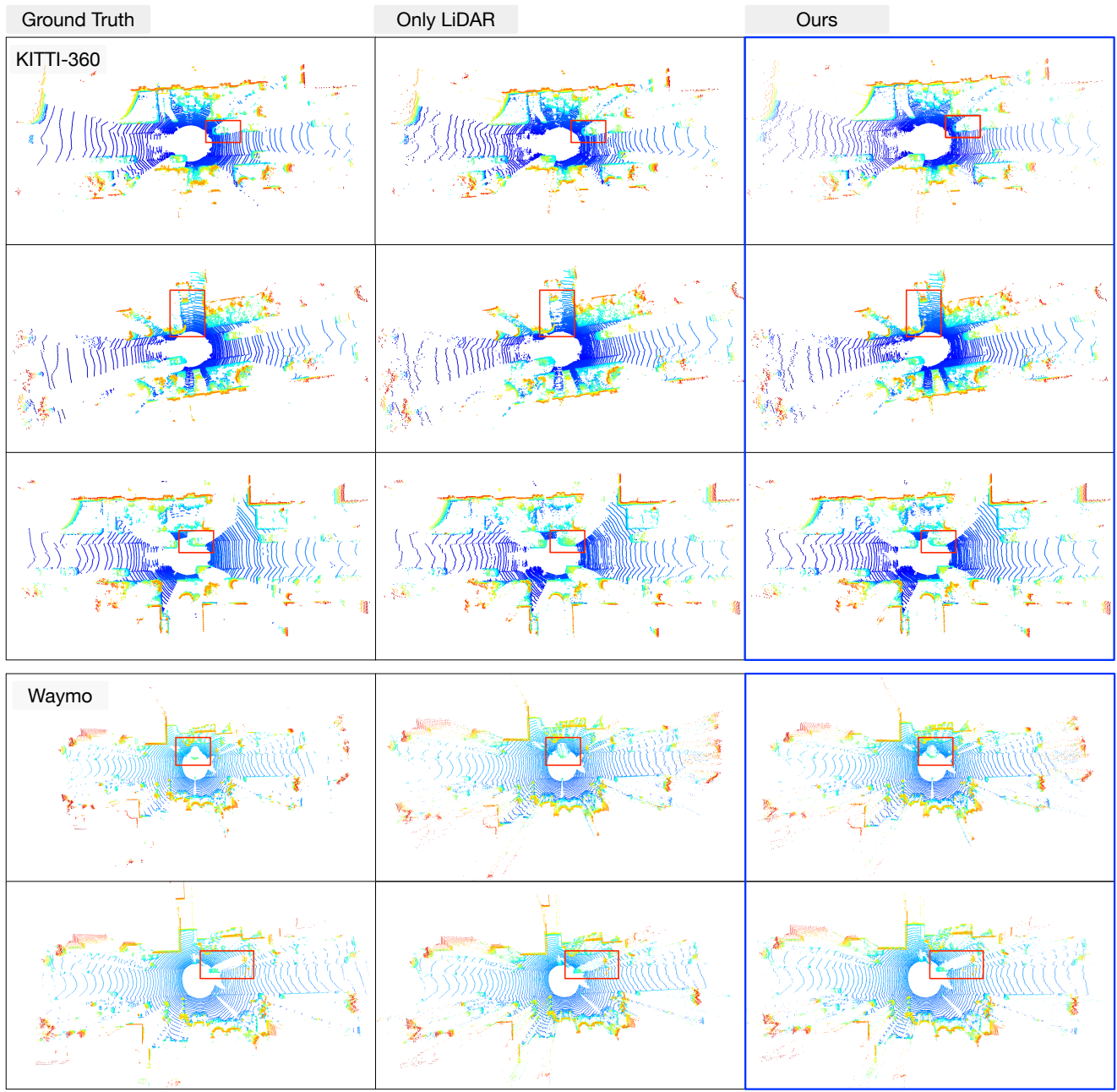}
    \caption{\textbf{Qualitative results of the LiDAR on KITTI-360 and Waymo datasets.} Our \name{} enhances information interactions between modalities and 
    semantic information from the camera aids the LiDAR in better converging to object boundaries (zoom-in for the best of views). Visualizing from a single perspective may not provide a comprehensive analysis of the LiDAR in 3D space. It's encouraged to try our code and models, and use 3D-view tools for a more comprehensive understanding of our method's superiority.
    }
    \label{fig:res_lidar}
\end{figure*}

\mypara{Qualitative results on KITTI-360 dataset.}
We provide more qualitative results on KITTI-360 dataset in \cref{fig:res_kitti} and \cref{fig:res_lidar}, which show the mutual benefits of our \name{}. The LiDAR modality significantly improves the learning of image and depth quality in the camera, while the semantic information from RGB assists the LiDAR in better converging to object boundaries.

\begin{figure*}[h]
    \centering
    \includegraphics[width=1\linewidth]{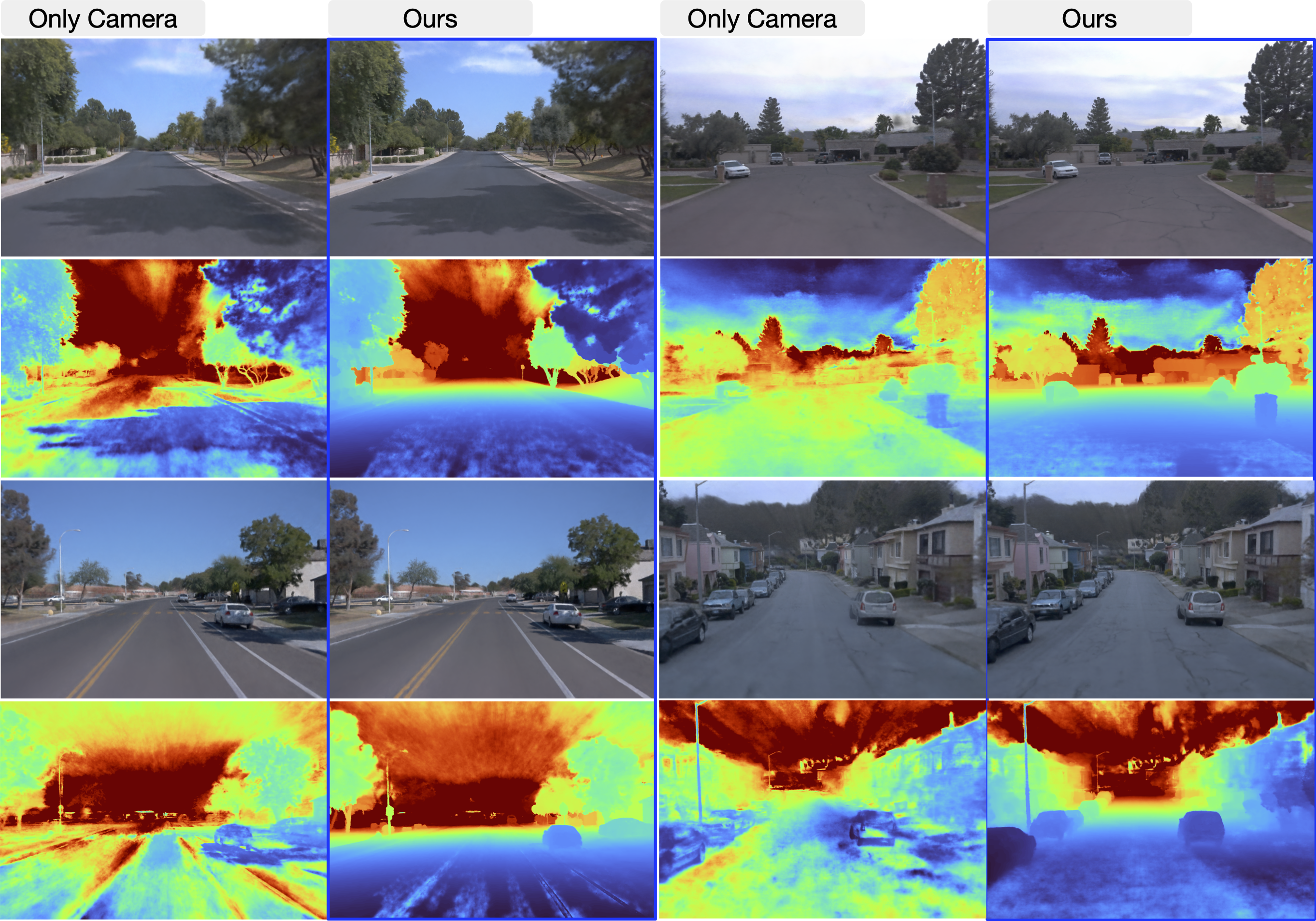}
    \caption{\textbf{Qualitative results of the camera on Waymo dataset.} Our \name{} enhances information interactions between modalities and improves image and depth quality in the camera using LiDAR information.}
    \label{fig:res_waymo}
\end{figure*}

\mypara{Qualitative results on Waymo dataset.}
We provide more qualitative results on the Waymo dataset in \cref{fig:res_waymo} and \cref{fig:res_lidar}, which demonstrate that the proposed \name{} significantly enhances the alignment and fusion between LiDAR and camera modalities, leading to more accurate join synthesis of LiDAR and camera novel views.

\mypara{Video demo.} In addition to the figures, we have attached a video demo in the supplementary materials, which consists of hundreds of frames that provide a more comprehensive evaluation of our proposed approach.

\begin{table*}[ht]
  \centering
  \small
  \caption{\textbf{Novel view synthesis on KITTI-360 dataset and Waymo dataset}. \name{} outperforms the baselines in all metrics.
  }
  \addtolength{\tabcolsep}{-4.65pt}
    \begin{tabularx}{0.95\linewidth}{l|c|ccc|ccc||ccc|ccc}
  \toprule
 \multirow{3}{*}{Method} &
 \multirow{3}{*}{M}  & \multicolumn{6}{c||}{KITTI-360 Dataset} & \multicolumn{6}{c}{Waymo Dataset}  \\
 \cmidrule(r){3-14} & & \multicolumn{3}{c|}{RGB Metric} & \multicolumn{3}{c||}{LiDAR Metric} & \multicolumn{3}{c|}{RGB Metric} & \multicolumn{3}{c}{LiDAR Metric} \\
       &    & PSNR$\uparrow$ & SSIM$\uparrow$ & LPIPS$\downarrow$ & C-D$\downarrow$ & F-score$\uparrow$  & MAE$\downarrow$ & PSNR$\uparrow$ & SSIM$\uparrow$ & LPIPS$\downarrow$ & C-D$\downarrow$ & F-score$\uparrow$  & MAE$\downarrow$\\
      \midrule

   \multicolumn{2}{l|}{Sequence} & \multicolumn{6}{l||}{\textit{Seq 1538--1601 }} & \multicolumn{6}{l}{\textit{seg11379226583756500423\_6230\_810\_6250\_810}} \\
       \midrule
    i-NGP~\cite{muller2022instantNGP}&C & 25.22 & 0.831 & 0.175 & --&--&-- &29.26 &0.825& 0.369 & --&--&--\\
     LiDAR-NeRF~\cite{tao2023lidarnerf}&L & --&--&-- & 0.088 &0.925&0.106 & --&--&-- & 0.216 & 0.853 &0.026\\
     \UniSim ~\cite{yang2023unisim} &LC & 22.92 \minus  &0.746 & 0.328 &0.083 \plus& 0.927 &0.103 &
     26.90 \minus  & 0.777& 0.399 &0.199 \plus &0.854 &0.027
     \\
     $\text{\UniSim ~\cite{yang2023unisim}}^{\triangledown}$ &LC & 25.25 \plus & 0.827& 0.184 &0.109 \minus &0.904 &0.102 &
     29.35 \plus &0.825& 0.367 &0.359 \minus & 0.761 &0.030
     \\
    \LC   \name{}  &LC & 25.67 \plus &0.837  &0.176  &0.079 \plus &0.930  &0.106 & 30.16 \plus &0.838& 0.331& 0.191 \plus& 0.863 &0.026\\

     \midrule
   \multicolumn{2}{l|}{Sequence} & \multicolumn{6}{l||}{\textit{Seq 1728--1791 }} & \multicolumn{6}{l}{\textit{seg10676267326664322837\_311\_180\_331\_180}} \\
       \midrule
    i-NGP~\cite{muller2022instantNGP}&C & 24.90 & 0.816 & 0.167  & --&--&-- & 29.52 &0.861 &0.299 & --&--&-- \\
     LiDAR-NeRF~\cite{tao2023lidarnerf}&L & --&--&--&0.107 & 0.895& 0.111 & --&--&--&   0.264 &0.841 &0.026\\
     $\text{\UniSim ~\cite{yang2023unisim}}^{\vartriangle}$  &LC & 23.86 \minus  & 0.782 & 0.228 & 0.097 \plus&  0.909 & 0.094 &
     26.47 \minus  & 0.808  & 0.341  & 0.254 \plus  & 0.848  & 0.026\\
   $\text{\UniSim ~\cite{yang2023unisim}}^{\triangledown}$ &LC &25.38 \plus &0.823& 0.167 &0.127 \minus& 0.891 &0.093 &
     29.54 \plus &0.862& 0.297& 0.612 \minus& 0.704 &0.039\\
    \LC   \name{}  &LC &25.43 \plus& 0.836 &0.148 &0.086 \plus& 0.913& 0.096& 30.27 \plus &0.873 &0.273 &0.228 \plus &0.859 &0.025\\

    \midrule
   \multicolumn{2}{l|}{Sequence} & \multicolumn{6}{l||}{\textit{Seq 1908--1971 }} & \multicolumn{6}{l}{\textit{seg17761959194352517553\_5448\_420\_5468\_420}} \\
       \midrule
      i-NGP~\cite{muller2022instantNGP}&C & 24.45 &0.787 &0.184 & --&--&-- & 28.20 &  0.830 & 0.372 & --&--&--\\
     LiDAR-NeRF~\cite{tao2023lidarnerf}&L & --&--&--&0.088 & 0.920&0.159 &--&--&--& 0.179& 0.885 &0.049\\
     $\text{\UniSim ~\cite{yang2023unisim}}^{\vartriangle}$  &LC & 23.54 \minus  &0.759 & 0.235 &0.087 \plus& 0.929 & 0.097 & 26.41 \minus & 0.789& 0.403 &0.173 \plus& 0.891 &0.049\\
    $\text{\UniSim ~\cite{yang2023unisim}}^{\triangledown}$ &LC & 24.65 \plus& 0.803 &0.172&0.111 \minus& 0.912&0.097& 28.33 \plus &0.830 &0.369 &0.227 \minus &0.840 & 0.052\\
    \LC   \name{}  &LC &25.20 \plus&0.816& 0.160 & 0.077 \plus&0.932&0.101   & 29.22 \plus & 0.841 & 0.327& 0.151 \plus& 0.896 &0.048\\

          \midrule
   \multicolumn{2}{l|}{Sequence} & \multicolumn{6}{l||}{\textit{Seq 3353--3416 }} & \multicolumn{6}{l}{\textit{seg1172406780360799916\_1660\_000\_1680\_000}} \\
       \midrule
    i-NGP~\cite{muller2022instantNGP}&C & 23.88 & 0.800& 0.199 & --&--&-- &28.30 &0.810 &0.480 & --&--&--\\
     LiDAR-NeRF~\cite{tao2023lidarnerf}&L & --&--&--& 0.094 &0.927 &0.112& --&--&-- &0.127 &0.908 &0.057\\
     $\text{\UniSim ~\cite{yang2023unisim}}^{\vartriangle}$  &LC & 22.90 \minus  & 0.746& 0.283 &0.091 \plus& 0.933 & 0.093&
     26.91 \minus& 0.778 & 0.526 &0.118 \plus& 0.919& 0.056\\
     $\text{\UniSim ~\cite{yang2023unisim}}^{\triangledown}$ &LC &24.51 \plus &0.797& 0.213  &0.110 \minus&0.919& 0.091 &
     28.72 \plus& 0.816& 0.463& 0.225 \minus &0.841 &0.058\\
    \LC   \name{}  &LC & 24.91 \plus & 0.815 &0.175& 0.082 \plus & 0.937& 0.096
    & 29.47 \plus &0.828& 0.427 &0.107 \plus & 0.921& 0.054\\

            \midrule
   \multicolumn{8}{l}{Average } & \multicolumn{6}{l}{Average }\\
       \midrule
    i-NGP~\cite{muller2022instantNGP}&C & 24.61 & 0.808 & 0.181 & --&--&-- &28.82	& 0.831	& 0.380	& --&--&--\\
     LiDAR-NeRF~\cite{tao2023lidarnerf}&L & --&--&--& 0.094 & 0.916 & 0.122
     & --&--&--&	0.197 &	0.871 & 0.040 \\
     $\text{\UniSim ~\cite{yang2023unisim}}^{\vartriangle}$  &LC & 23.30 \minus  & 0.758 & 0.268 & 0.090 \plus& 0.924 & 0.097
     & 26.67 \minus 	&0.788	&0.417&	0.186 \plus&	0.878	&0.039 \\
    $\text{\UniSim ~\cite{yang2023unisim}}^{\triangledown}$ &LC & 24.94 \plus&	0.812&	0.184&	0.114 \minus&0.906&	0.095
     & 28.98 \plus&	0.833&	0.374 &	0.355 \minus&	0.786 &	0.045 \\
   \LC   \name{}  &LC & 25.31 \plus&	0.826	&0.164&	0.081 \plus&	0.928	&0.099 &
   29.78 \plus	&0.845 &	0.339&	0.169 \plus &	0.885&	0.038\\

      \bottomrule
      \multicolumn{14}{l}{\scriptsize{M, L, C denotes modality, LiDAR, camera respectively. $\vartriangle$ and $\triangledown$ represent tuning parameters towards LiDAR and camera modality respectively.}}
    \end{tabularx}
  \label{tab:all_res_full}
\end{table*}

\end{document}